# Fully-automated sleep staging: multicenter validation of a generalizable deep neural network for Parkinson's disease and isolated REM sleep behavior disorder

## Authors

Jesper Strøm[1†], Casper Skjærbæk[2,3†*], Natasha Becker Bertelsen[2,3], Steffen Torpe Simonsen[1], Niels Okkels[4], David Bertram[5,6,7], Sinah Röttgen[8,9,10], Konstantin Kufer[10,11], Kaare B. Mikkelsen[1], Marit Otto[4], Poul Jørgen Jennum[12], Per Borghammer[2,3], Michael Sommerauer[8,9,10,11], Preben Kidmose[1]

[1]Department of Electrical and Computer Engineering, Aarhus University, Aarhus, Denmark
[2]Department of Nuclear Medicine, Aarhus University Hospital, Aarhus, Denmark
[3]Lundbeck Foundation Parkinson's Disease Research Center (PACE), Aarhus University, Aarhus, Denmark
[4]Department of Neurology, Aarhus University Hospital, Aarhus, Denmark.
[5]Faculty of Mathematics and Natural Sciences, University of Cologne, Cologne, Germany
[6]Institute for Biomedical Informatics, Faculty of Medicine and University Hospital Cologne, Cologne, Germany
[7]Center for Molecular Medicine Cologne (CMMC), Faculty of Medicine and University Hospital Cologne, Cologne, Germany
[8]Cognitive Neuroscience, Institute for Neuroscience and Medicine, INM-3, Research Center Juelich, Germany.
[9]Department of Neurology, Faculty of Medicine and University Hospital Cologne, University of Cologne, Germany.
[10]Center of Neurology, Department of Parkinson, Sleep and Movement Disorders, University Hospital Bonn, University of Bonn, Germany.
[11]German Center for Neurodegenerative Diseases (DZNE), Bonn, Germany.
[12]Danish Center for Sleep Medicine, Glostrup University Hospital, Glostrup, Denmark.

*Corresponding author: cas@clin.au.dk
†These authors made equal contributions

**Abstract**

Isolated REM sleep behavior disorder (iRBD) is a key prodromal marker of Parkinson's disease (PD), and video-polysomnography (vPSG) remains the diagnostic gold standard. However, manual sleep staging is labor-intensive and particularly challenging in neurodegenerative diseases due to EEG abnormalities and fragmented sleep, making PSG assessments a bottleneck for deploying new RBD screening technologies at scale.

We adapted U-Sleep, a deep neural network, for generalizable sleep staging in PD and iRBD. A pretrained U-Sleep model, based on a large publicly available, multisite non-neurodegenerative dataset (PUB; 19,236 PSGs across 12 sites), was fine-tuned on research datasets from two centers (Lundbeck Foundation Parkinson's Disease Research Center (PACE) and the Cologne-Bonn Cohort (CBC); 112 PD, 138 iRBD, 89 age-matched controls). The resulting model was evaluated on an independent hold-out dataset of clinical PSG recordings from the Danish Center for Sleep Medicine (DCSM; 81 PD, 36 iRBD, 87 sleep-clinic controls). Linear mixed-effects models were used to test predictors of Cohen's κ, including U-Sleep confidence, polysomnographic, and demographic variables. A subset of PSGs with low agreement between the human rater and the model (κ < 0.6) was re-scored by a second blinded human rater to identify sources of disagreement. Finally, we applied confidence-based thresholds to optimize REM sleep staging.

The pretrained model achieved mean κ = 0.81 in PUB, but κ = 0.66 when applied directly to PACE/CBC. By fine-tuning the model, we developed a generalized model with κ = 0.74 on PACE/CBC (p < 0.001 vs. the pretrained model) and markedly improved F1 across all stages (p < 0.001). On the DCSM hold-out dataset, mean and median κ increased from 0.60 to 0.64 (p < 0.001) and 0.64 to 0.69 (p < 0.001), respectively. Site-specific fine-tuning produced no or only marginal improvements. U-Sleep confidence was the only independent predictor of per-night Cohen's κ in models restricted to information known before human scoring (p < 0.001). In the interrater study, PSGs with low agreement between the model and the initial scorer showed similarly low agreement between human scorers, suggesting that these recordings are inherently challenging to score. Applying a confidence threshold increased the proportion of correctly identified REM sleep epochs from 85% to 95.5%, while preserving sufficient (> 5 min) REM sleep for 95% of subjects.

This fine-tuned U-Sleep model is made publicly available and achieves sleep-staging performance comparable to human interrater agreement across cohorts of PD, iRBD, and clinically relevant controls, including an unseen clinical site. Model-derived confidence provides a practical tool for predicting agreement and target REM sleep epochs, enabling scalable, standardized PSG analysis for clinical and research settings.

## Introduction

Over the past 25 years, the global burden of Parkinson's disease (PD) has more than doubled, and PD is now the fastest-growing neurological disorder in terms of prevalence, disability, and deaths[1], with further increases anticipated through 2050[2]. PD belongs to the group of α-synucleinopathies together with dementia with Lewy bodies (DLB), multiple system atrophy (MSA), and isolated REM sleep behavior disorder (iRBD). RBD is characterized by dream-enactment behaviors, such as shouting, kicking, and complex movements during vivid dreaming in REM sleep. Although these symptoms may appear benign, iRBD in most cases reflects underlying α-synucleinopathy affecting the brainstem[3]. As the strongest prodromal marker of PD, the likelihood of later PD is increased by approximately 130-fold[4]. After 12–14 years of follow-up, 73–90% of individuals with iRBD develop an overt α-synucleinopathy[5,6]. As a result, iRBD is fast becoming a key entry point for α-synucleinopathy trials[7].

Extensive screening efforts seek to identify iRBD before the onset of overt motor or cognitive symptoms. These include community-based questionnaire screening[8-10], device-based approaches using actigraphy and wearable EEG/EM[11-14], and, more recently, assays of α-synuclein aggregation in blood or cerebrospinal fluid[15,16]. But, none currently provides sufficient sensitivity and specificity to replace video-polysomnography (vPSG) as the diagnostic gold standard for REM sleep behavior disorder (RBD). Guideline-based evaluation relies on at least one vPSG, typically performed in-hospital, to document REM sleep without atonia (RSWA) and dream enactment behaviours[17], and requires multichannel EEG, electrooculography (EOG), multiple electromyography (EMG) channels, cardiorespiratory monitoring, and synchronized video recording. Sleep stages are then assigned manually to consecutive 30-s epochs (≈1,000 per night) in accordance with the AASM Scoring Manual[18]. As questionnaire- and biomarker-based screening scales up, labor-intensive confirmatory vPSG assessments become a bottleneck to the deployment of new technologies for RBD screening.

Sleep staging in neurodegenerative disease requires substantial expertise at each sleep center to provide consistent results. In healthy adults, interrater agreement when applying AASM rules is substantial, with Cohen's κ values around 0.76[19]. In PD and related disorders, pathological changes in sleep EEG are common, including EEG slowing during wakefulness[20,21], attenuation or loss of N2 hallmarks such as sleep spindles[22,23], and reduced or atypical rapid eye movements during REM sleep[24]. These deviations from canonical sleep architecture undermine the standard AASM rules and reduce interrater agreement, with Cohen's κ ≈ 0.61 reported in the largest PD study[25] and slightly higher values in a smaller study[26]. In neurodegenerative disease, strict application of AASM rules may, in some cases, be impractical or impossible, for example, in patients without definite rapid eye movements or with absence of REM atonia. To address this, RBD-specific staging recommendations for REM sleep and modified criteria for sleep in neurodegenerative disease have been proposed[17,23,27]. While such adaptations may improve consistency within a center, they also increase complexity and introduce site-specific practices and inconsistent staging, which complicate the analysis of human-scored data and limit the performance of supervised models trained on these annotations.

Automated sleep staging could bypass such variability and alleviate the current bottleneck of manual sleep staging. Deep neural network models have advanced substantially in recent years, now achieving agreement with human raters comparable to human–human interrater agreement in non-neurodegenerative cohorts[28,29]. U-Sleep, a fully convolutional neural network, robustly reproduces staging across diverse non-neurodegenerative populations when trained on large multisite datasets[28,30]. Yet models are typically trained on datasets dominated by healthy adults, and none have been adapted or validated for use in neurodegenerative diseases. Domain shift due to pathological EEG, fragmented sleep, comorbid sleep disorders, and inconsistent human labels may substantially degrade performance. This is supported by prior work applying a random-forest classifier to PSG from individuals with iRBD, in which agreement was substantial in elderly controls (κ = 0.73) but dropped markedly in iRBD (κ = 0.54), with less than half of manually scored REM sleep correctly identified[31]. This pattern underscores that models performing well in healthy cohorts may fail in challenging populations, including iRBD.

A prerequisite for scalable use of PSG in neurodegenerative disease is a robust, generalizable sleep-staging model that maintains near-human agreement across centers, disease stages, and comorbid sleep disorders, and that provides calibrated measures of confidence to guide manual review. Our objective was to adapt and validate the U-Sleep model for application in PD, iRBD, and clinically relevant controls across multiple sites. We fine-tuned a pretrained U-Sleep model, initially trained on large, multisite, non-neurodegenerative datasets, using high-quality research PSGs from two research centers, comprising individuals with iRBD, PD without co-morbid RBD (PD$^{-RBD}$), PD with RBD (PD$^{+RBD}$), and age-matched controls recruited primarily through questionnaire-based screening initiatives[8,9] and a sleep clinic. Importantly, we did not exclude co-morbid sleep disorders such as OSA and insomnia, to reflect populations encountered in clinical and screening settings.

To inform clinical deployment, we then addressed five key questions: (i) how much performance degrades when the pretrained model is applied directly to neurodegenerative datasets and to what extent finetuning restores agreement, including whether a single generalized model can replace site-specific models; (ii) how well the generalized model performs in an independent hold-out cohort; (iii) which factors (e.g. age, apnea–hypopnea index, U-Sleep confidence) determine agreement; (iv) whether, in challenging recordings, a blinded second human scorer agrees more closely with the model or the original scorer, thereby revealing signs of overfitting or systematic deviations (v) whether model confidence can be used to develop purpose-specific adaptations, most importantly, optimized REM sleep detection using thresholds preserving sufficient REM sleep duration for downstream clinical review or automated RSWA assessment in PD and iRBD.

## Results

We first replicated the original U-Sleep model on a large, publicly available multisite dataset of healthy individuals, children, and patients with obstructive sleep apnea (PUB; 19,236 PSG recordings across 12 sites; Table S1), achieving a validation Cohen's κ of 0.813 and a per-night κ of 0.818 on the test set. This *Pretrained Model* was then fine-tuned on two high-quality research datasets (PACE and CBC) comprising patients with neurodegenerative disorders (112 PD and 138 iRBD) and age-matched controls (n = 89; Table 4), resulting in a *Generalized Model*. Additionally, *Site-Specific Models* were trained on PACE and CBC individually. Finally, the Generalized Model was evaluated on an independent clinical hold-out dataset from the Danish Centre of Sleep Medicine (DCSM), enriched with patients with neurodegenerative disorders (81 PD and 36 iRBD) and sleep-clinic patients without neurodegenerative disease (n = 87). The Generalized Model is available on our GitHub repository (https://github.com/jesperstroem/U-Sleep-for-RBD-PD).

### Sleep staging performance
#### Pretrained multicenter model
A total of 19,236 PSG recordings were included in PUB (Table S1). Per-night κ and stage-wise F1-scores for the reproduced Pretrained Model are presented in Table S2. As we used a different data stratification and computed κ and F1 on a per-night rather than a per-dataset basis, these scores are not directly comparable to those reported by Perslev et al.[28] To demonstrate comparable performance, per-dataset F1-scores for both models are reported in Table S3.

#### Fine-tuned model for neurodegenerative diseases
The distribution of per-night κ scores between the model and the human scorer is shown in Figure 1 for all three model types: the Pretrained Model, the Generalized Model fine-tuned on the PACE and CBC datasets with neurodegenerative patients, and the Site-Specific Models trained on a single center, all evaluated using 10-fold cross-validation. Accuracy, κ, and F1-scores for each model are provided in Table 1, and the metrics for each dataset are provided in Table S5.

Across 339 research recordings in total (PACE = 134, CBC = 205, Table S4), average per-night κ increased significantly from the Pretrained Model to the Generalized Model at both sites ($p < 0.001$), and F1-scores increased across all sleep stages ($p < 0.001$). Site-Specific Models yielded a statistically significant improvement only for CBC (PACE: $p = 0.076$; CBC: $p < 0.001$). When combining stages N1 and N2 into *Light Sleep*, the overall κ of all models increased significantly ($p < 0.001$), and for the Generalized Model, the F1-score for Light Sleep reached 0.88.

For the unseen "hold-out" DCSM dataset (Figure 1, right panel), the Pretrained Model without finetuning produced a mean per-night κ of 0.60, which improved significantly to 0.64 with the Generalized Model ($p < 0.001$). Median per-night κ also improved from κ = 0.64 to κ = 0.69 ($p < 0.001$). Although neither the Pretrained nor the Generalized Model was trained on DCSM data, finetuning specifically to the DCSM site did not yield significant improvements in median κ or stage-wise F1 scores compared to using the Generalized Model (κ = 0.69 vs. 0.70, $p = 0.42$).

Finetuning the Pretrained Model resulted in significant improvements across all sleep stages compared to the Pretrained Model on PACE and CBC (p < 0.001 for all sleep stages). Site-Specific Models have significantly better staging of Wake, N1 and REM (W: p = 0.013, N1: p = 0.009, N2: p = 0.838, N3: p = 0.467, R: p = 0.005).

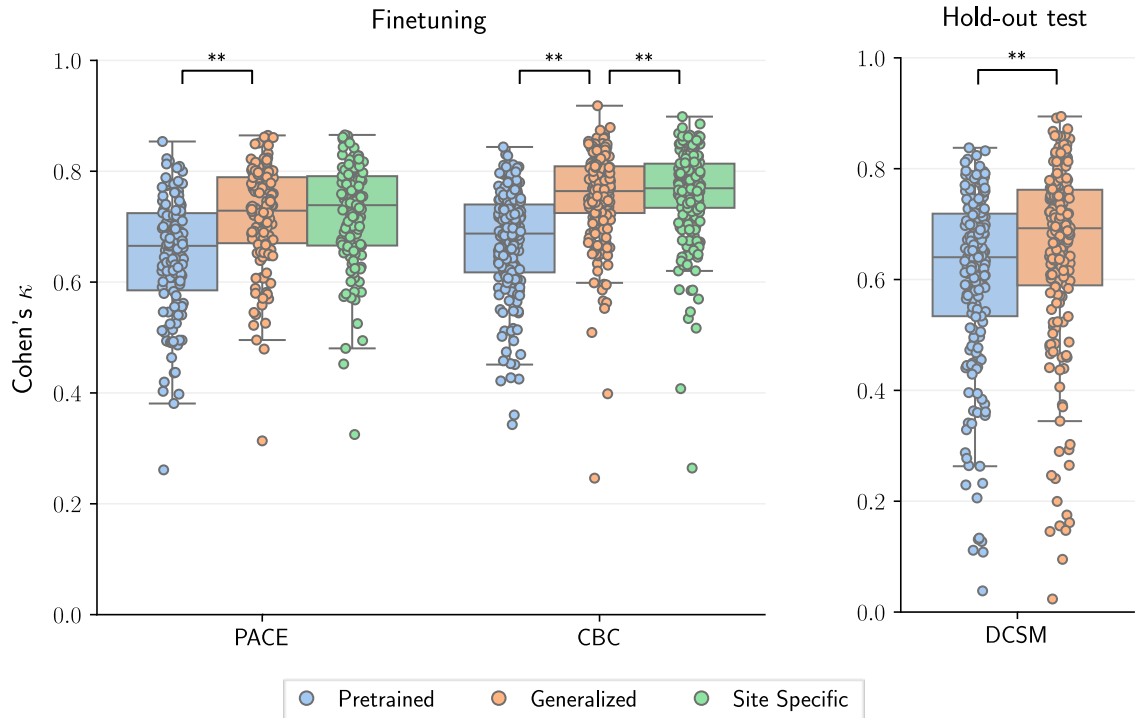

*Figure 1* – *Distribution of per-night κ values for each model when applied to datasets individually.* **Left panel:** *The Pretrained Model, similar to the model in Perslev et al., achieved median κ values of 0.67 and 0.69 for PACE and CBC, respectively, and improved significantly upon fine-tuning. Only in CBC, is the Site-Specific Model marginally better than the Generalized Model.* **Right panel:** *The DCSM dataset was used as the hold-out set, and median κ increased from 0.64 to 0.69 using the Generalized Model compared to the Pretrained Model. *Indicates significant improvements (*: p < 0.05, **: p < 0.001)*

### Clinical subgroups

The datasets comprise four clinical subgroups, including PD patients without RBD (PD$^{-RBD}$), PD patients with RBD (PD$^{+RBD}$), iRBD, and controls, as shown in Table 4. Controls in PACE were slightly older than patients, although not significant (p = 0.06), while controls in CBC were significantly younger than patients (p = 0.01). For both PACE and CBC, controls tended to have higher apnea-hypopnea index, although only significant for CBC (PACE: p = 0.10, CBC: p < 0.001, respectively). In contrast, controls from DCSM were younger and had lower AHI than patients (p < 0.001, p = 0.011, respectively).

For PACE and CBC, no significant differences in median per-night κ were observed across clinical subgroups, except for the Pretrained Model on iRBD and PD$^{-RBD}$ (p < 0.001), as shown in Figure 2 (left). A non-significant difference was observed for the Site-Specific Model between PD$^{-RBD}$ and PD$^{+RBD}$. As shown in Figure 2 (right) the Generalized Model showed significant differences between groups when comparing stage-wise F1-scores for N1 and N3 in PD patients, with PD$^{+RBD}$ generally having the lowest F1 score (p < 0.001). No stage-wise differences were found for W, N2 and R, except in W between iRBD and PD$^{-RBD}$ (p < 0.001) and in R between iRBD and PD$^{-RBD}$ (p = 0.048). Means and standard deviations are provided in Table S6.

In DCSM, the Generalized Model showed significant differences in median κ between controls and all subgroups (Control: κ = 0.75, iRBD: κ = 0.65, PD$^{-RBD}$: κ= 0.59, PD$^{+RBD}$: κ= 0.64, p < 0.001). Applying a Site-Specific Model for DCSM showed only non-significant improvements within subgroups compared to the Generalized Model (Control: p = 0.45, iRBD: p = 0.84, PD$^{-RBD}$: p = 0.12, PD$^{+RBD}$: p = 1.0).

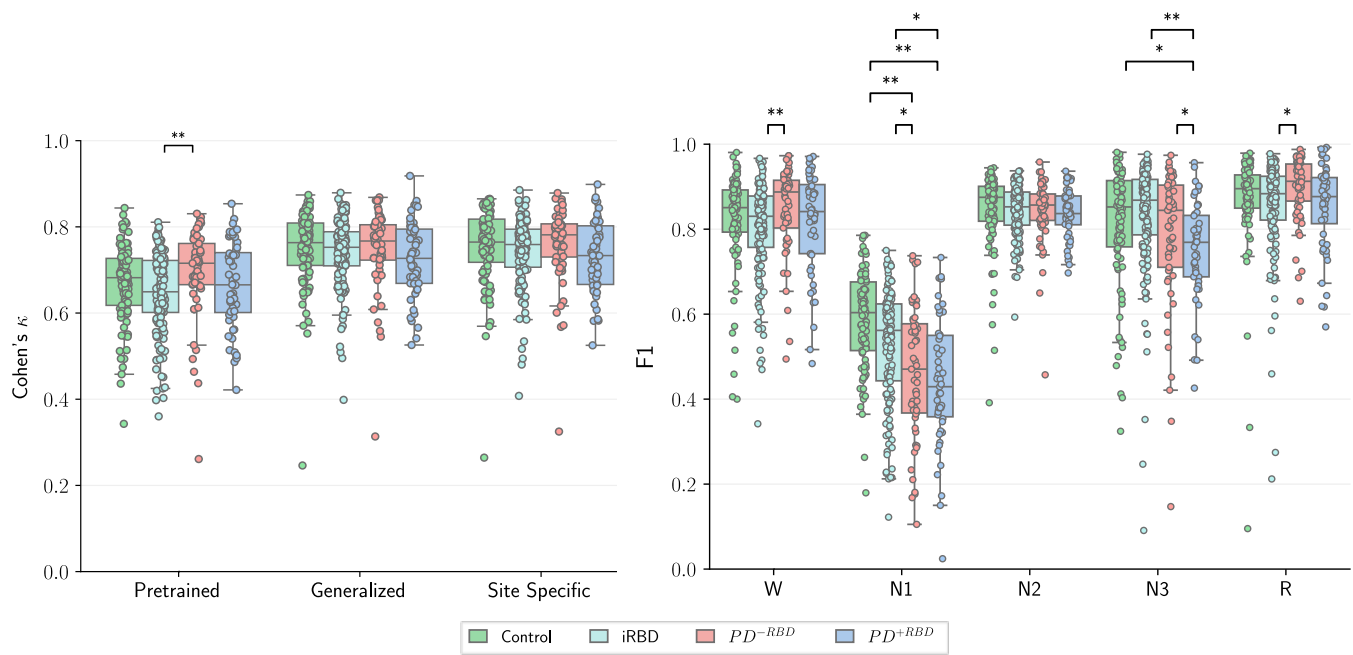

*Figure 2 – Left:* *Distribution of per-night κ scores for the models across clinical subgroups in PACE and CBC. A significant difference was found between iRBD and PD$^{-RBD}$ for the Pretrained Model.* *Right:* *Stage-wise performance across each clinical subgroup in PACE and CBC for the Generalized Model. *Indicates significant improvements (*: p < 0.05, **: p < 0.001).*

*Sleep architecture measures*

Sleep architecture measures for PACE and CBC are shown in Table 1 and Figure S3. For the Pretrained Model, median values differed significantly from human scoring across all parameters (p = 0.001 for SEFF; p < 0.001 for the other parameters), except for SOL (p = 1.0), WASO (p = 1.0), and PREM (p = 0.16). The Generalized and Site-Specific Models also differ on most parameters, most notably on the number of stage changes, which is markedly lower across all U-sleep models (p < 0.001).

|  | Metric | Human | Pretrained | Generalized | Site Specific |
|---|---|---|---|---|---|
|  | Accuracy |  | 0.76 ± 0.07 | 0.82 ± 0.06 | 0.82 ± 0.06 |
|  | Cohen's κ |  | 0.66 ±0.10 | 0.74 ± 0.09 | 0.75 ± 0.09 |
| **F1 Scores** | W |  | 0.74 ± 0.13 | 0.82 ± 0.11 | 0.82 ± 0.11 |
|  | N1 |  | 0.39 ± 0.14 | 0.52 ± 0.15 | 0.53 ± 0.16 |
|  | N2 |  | 0.81 ± 0.08 | 0.84 ± 0.07 | 0.84 ± 0.07 |
|  | N3 |  | 0.74 ± 0.17 | 0.80 ± 0.14 | 0.80 ± 0.14 |
|  | R |  | 0.80 ± 0.16 | 0.87 ± 0.11 | 0.87 ± 0.11 |
|  | Average |  | 0.70 ± 0.07 | 0.77 ± 0.07 | 0.77 ± 0.07 |
| **Sleep architecture** | Total Sleep Time, TST (m) | 396.27 ± 67.89 | 401.76 ± 64.24 ** | 396.64 ± 64.63 | 396.61 ± 63.51 |
|  | Sleep Efficiency, SEFF (%) | 80.7 ± 10.84 | 81.83 ± 9.88 * | 80.78 ± 10.08 | 80.83 ± 10.23 |
|  | Sleep Onset Latency, SOL (m) | 11.15 ± 12.91 | 12.08 ± 14.73 | 12.41 ± 13.94 | 12.57 ± 14.66 |
|  | REM Latency, REMLAT (m) | 124.44 ± 81.11 | 124.37 ± 84.99 ** | 118.05 ± 76.22 ** | 116.6 ± 74.79 ** |
|  | Fraction of N1, PN1 (%) | 16.2 ± 6.67 | 16.14 ± 7.31 ** | 17.37 ± 6.99 ** | 17.17 ± 6.91 ** |
|  | Fraction of N2, PN2 (%) | 17.81 ± 11.31 | 12.27 ± 8.86 ** | 14.38 ± 8.91 ** | 14.36 ± 9.75 * |

| | | | | |
|---|---|---|---|---|
| Fraction of N3, PN3 (%) | 50.25 ± 10.23 | 57.33 ± 10.13 ** | 51.69 ± 9.57 * | 51.91 ± 10.2 * |
| Fraction of R, PREM (%) | 15.75 ± 7.62 | 14.26 ± 8.05 | 16.56 ± 7.45 ** | 16.55 ± 7.45 ** |
| Wake After Sleep Onset, WASO (m) | 80.55 ± 55.02 | 79.16 ± 50.77 | 83.94 ± 52.29 ** | 83.82 ± 52.91 |
| Stage Changes (n) | 145.87 ± 53.99 | 108.41 ± 41.06 ** | 128.09 ± 47.83 ** | 129.22 ± 48.85 ** |

*Table 1* – Mean and standard deviation of κ, F1 scores, and sleep architecture measures for each model evaluated on PACE and CBC. Stars indicate a significant difference from human scorings (** $p < 0.001$, * $p < 0.05$).

### Latency and length of REM sleep and N2 periods

Across all recordings in PACE and CBC, 1017 overlapping R-periods and 3271 overlapping N2-periods were identified between human labels and the Generalized Model's sleep staging. The mean offset was −0.61 epochs for R and −0.54 epochs for N2, indicating that the model, on average, starts classifying both stages earlier than human scorers ($p < 0.001$). The difference in length was 1.18 epochs for R and 0.84 epochs for N2, indicating that the model scores longer consecutive periods of both stages than human raters ($p < 0.001$), as shown in Figure 3.

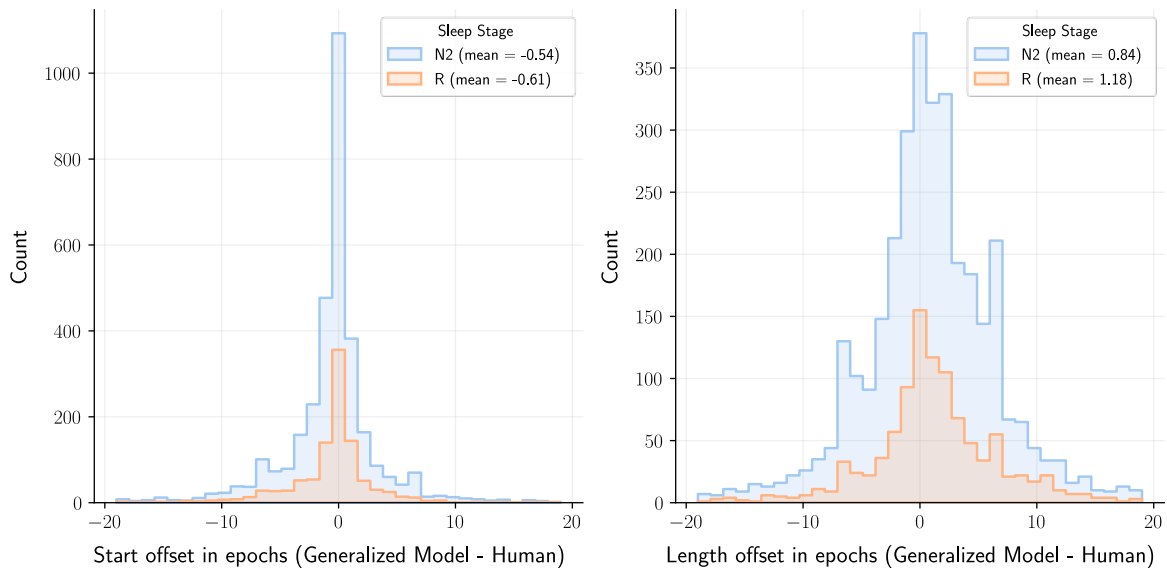

*Figure 3* – Distributions of start and length offsets for the identified REM sleep and N2 periods when predicted by the Generalized Model compared to human raters. N2 and R periods begin slightly earlier in the predicted sleep stages than when staged by human raters, and they continue for longer. This supports the notion that the model does not rely solely on rapid eye movements to initiate REM sleep but has adapted to accommodate the rules for "staging back" REM sleep in contiguous epochs preceding definite R epochs.

### Explaining sleep staging performance
#### Interrater disagreement

Thirty-one PSG recordings were included in a post hoc interrater study involving a new scorer, blinded to the U-Sleep predictions and the original hypnogram. Of these, 21 recordings were selected because κ between the Site-Specific Model and the original scoring was < 0.6, and 10 additional recordings with κ > 0.6 were selected. The original scorer (Rater 1, R1) and U-Sleep (Model) had a mean κ of 0.59, while R1 and the blinded scorer (Rater 2, R2) had a mean κ of 0.54. Eight of the chosen PSG recordings had a higher κ between R1 and R2 than between R1 and the model, as shown in the left panel of Figure 4. Mean κ between R2 and the model was 0.58, indicating that the model agreed at least as well with each rater as they did with each other.

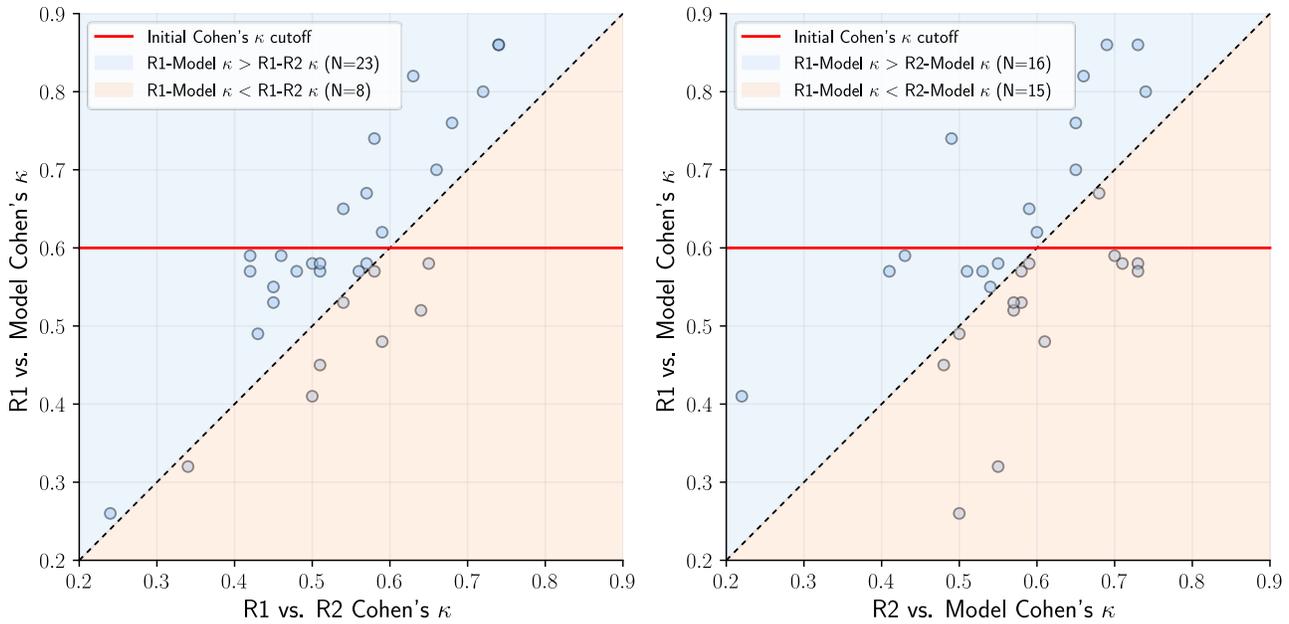

***Figure 4*** - *Scatterplots of κ values in the interrater study. The red line indicates the 0.6 κ cutoff for rescoring records. The numbers in the left and right panels indicate the number of data points within the areas defined by red and green colors.* ***Left:*** *Rater 1 (R1) vs. model and Rater 1 (R1) vs. Rater 2 (R2). Overall, better κ values were achieved for R1-R2 on only 8/31 PSG recordings, as shown in the lower-right area, suggesting that the agreement between R1 and the model is on par with or exceeds the human interrater agreement.* ***Right:*** *R1 vs. Model and R2 vs. Model. When the initial κ between R1 and the model is low, the κ between R2 and the model is often higher, and vice versa. This suggests a "regression towards the mean" effect.*

Table 2 summarizes accuracy, κ, and F1 scores for all rater pairs. Consensus κ was defined as the κ between the model scoring and the epochs in which R1 and R2 agree and reached a mean of 0.74 across all recordings, with κ = 0.69 and κ = 0.84 for recordings with initial κ below or above 0.6, respectively. Across all recordings, R1 and the Model showed higher agreement on N3 (p < 0.001) than R1 and R2. For those with low initial κ, R1 and R2 showed higher agreement than R2 and the Model only for N1, while R2 and the model reached greater agreement on W than either R1 vs. Model or R1 vs. R2 (p < 0.001). When initial κ was high, R1 and the model showed significantly higher agreement on average and for N1 and N3 than either R1 vs. R2 or R2 vs. Model (p < 0.001). Across all 31 recordings, R1 and R2 show the lowest agreement, as measured by κ, but this difference is significant only for R1 vs. Model (p = 0.002)

| | | | | | F1 | | | | | |
|---|---|---|---|---|---|---|---|---|---|---|
| Type | Relation | Accuracy | Cohen's κ | Human Consensus Cohen's κ | Wake | N1 | N2 | N3 | R | Average |
| Initial κ < 0.6 | R1 vs. Model | 0.67±0.06 | 0.52±0.09 | 0.69±0.13 | 0.66±0.16 | **0.38±0.17*** | 0.72±0.11 | **0.58±0.27*** | 0.71±0.26 | 0.61±0.10 |
| | R1 vs. R2 | 0.66±0.07 | 0.49±0.09 | | 0.65±0.14 | 0.47±0.15 | 0.69±0.15 | 0.36±0.26 | 0.70±0.24 | 0.58±0.07 |
| | R2 vs. Model | 0.69±0.10 | 0.55±0.11 | | **0.77±0.11*** | 0.45±0.15 | 0.70±0.18 | 0.39±0.27 | 0.70±0.27 | 0.61±0.10 |
| Initial κ > 0.6 | R1 vs. Model | **0.82±0.06*** | **0.75±0.08*** | 0.84±0.06 | 0.83±0.11 | 0.51±0.12 | **0.86±0.03*** | **0.81±0.08*** | 0.84±0.08 | **0.77±0.06*** |
| | R1 vs. R2 | 0.75±0.05 | 0.64±0.07 | | 0.80±0.10 | 0.49±0.09 | 0.78±0.06 | 0.52±0.24 | 0.83±0.13 | 0.69±0.07 |
| | R2 vs. Model | 0.75±0.05 | 0.65±0.07 | | 0.82±0.11 | 0.48±0.11 | 0.78±0.07 | 0.54±0.26 | 0.81±0.15 | 0.69±0.07 |

***Table 2*** - *Mean and standard deviation for accuracy, κ, and F1 scores for all rater relations – Rater 1 (R1), Rater 2 (R2), and model. Values in bold accompanied with * indicates a statistically significant differences from the two other rater relations within the type of recording (divided by initial κ value) (p < 0.05).*

*Predicting Cohen's κ*

Linear mixed-effects models were used to identify variables associated with per-night κ for the Generalized Model on PACE and CBC. The results are shown in Table 3. Model A included only subject-level information available before clinical PSG and identified age as the only significant predictor of κ. Model B included information available after routine PSG evaluation and identified AHI as an additional predictor, with a slope similar to that of age. Model C included only variables available before PSG or obtained via automatic sleep staging with U-Sleep; in this model, U-Sleep confidence was the only statistically significant predictor, with κ expected to increase by 0.011 for each 1%-point increase in U-Sleep confidence. The effect of age became non-significant. The impact of each variable on overall confidence is illustrated in Figure S4. In 10-fold cross-validation, Model C yielded a mean absolute error of 0.052 in κ, and the residuals are shown in Figure S5. The residuals are evenly distributed around 0, and the spread is not correlated with the predicted value, indicating that the mixed linear effect model provides a sufficient fit.

| | Model A: Demographics | | | Model B: Demographics + Sleep Disorders | | | Model C: Demographics + U-Sleep | | |
|---|---|---|---|---|---|---|---|---|---|
| Parameter | Coef. | z | p | Coef. | z | p | Coef. | z | p |
| Intercept | 0.976 | 18.853 | **<0.001*** | 0.912 | 17.827 | **<0.001*** | -0.100 | -0.918 | 0.359 |
| Age | -0.003 | -5.498 | **<0.001*** | -0.002 | -3.827 | **<0.001*** | -0.001 | -1.818 | 0.069 |
| Sex (M) | -0.014 | -1.279 | 0.201 | -0.013 | -1.199 | 0.230 | 0.010 | 1.078 | 0.281 |
| BMI | -0.002 | -1.290 | 0.197 | -0.000 | -0.143 | 0.887 | -0.000 | -0.330 | 0.741 |
| PD | 0.009 | 0.772 | 0.440 | 0.005 | 0.478 | 0.632 | -0.001 | -0.076 | 0.940 |
| AHI | | | | -0.002 | -4.329 | **<0.001*** | | | |
| RBD | | | | -0.015 | -1.588 | 0.112 | | | |
| PN1 % | | | | | | | -0.001 | -1.008 | 0.313 |
| STAGEC # | | | | | | | -0.000 | -0.232 | 0.817 |
| U-Sleep Confidence % | | | | | | | 0.011 | 10.836 | **<0.001*** |

*Table 3* - Overview of regression model parameters, their z scores, and p-values. The grey areas represent parameters that the given model has not been fitted to. PN1 % is the percentage of N1 sleep, STAGEC # is the total number of stage changes in the hypnogram and U-Sleep Confidence % is the percentage of mean confidence of the U-Sleep model when predicting a given sleep-stage.

*Correlation between model confidence and κ*

Visualizations of the epoch-wise distributions of model confidence are exemplified as hypodensity plots in Figure 5 for a recording with high average confidence (left, mean confidence = 0.8785) and low average confidence (right, mean confidence = 0.6415). For the Generalized Model, the mean confidence across all recordings was 0.80, with W having the highest confidence and N1 the lowest (W = 0.83, N1 = 0.58), as shown in Figure 6, right. A linear relationship between per-night overall confidence and κ-value was observed ($R^2$ = 0.40, p < 0.001, Figure 6, left), and a linear relationship between R-stage confidence and R F1 was also found (R2 = 0.45, p < 0.001).

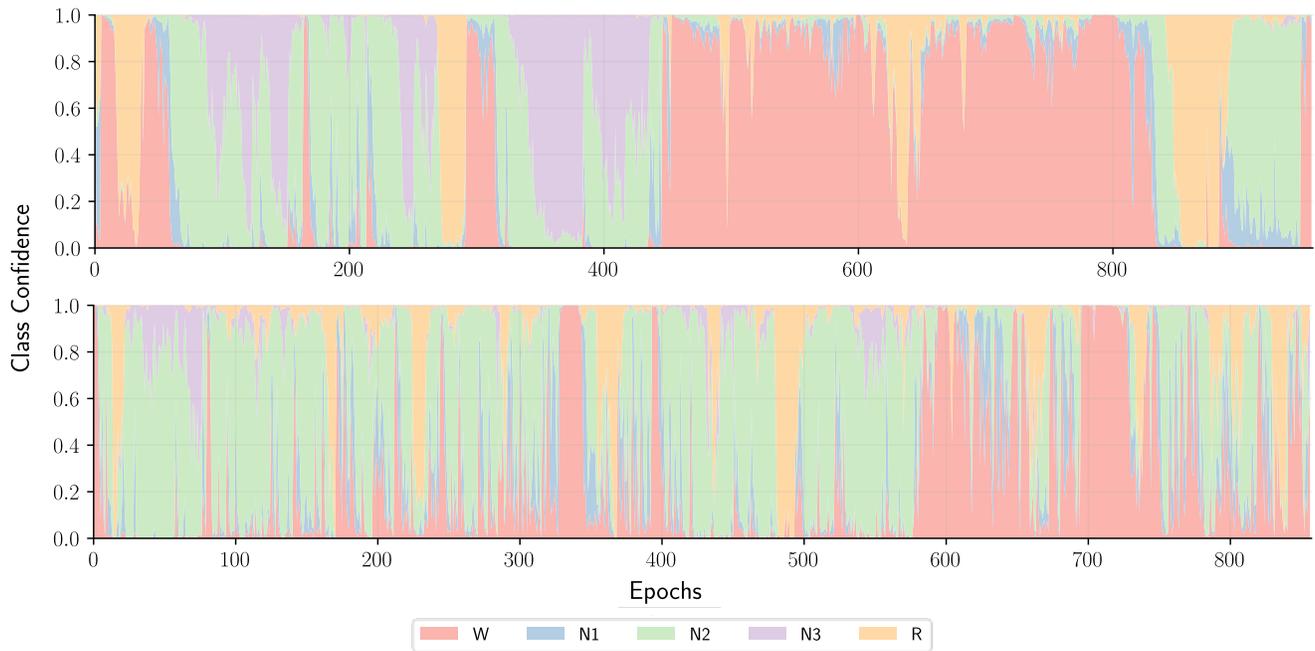

***Figure 5** – Confidence estimates. Examples of hypnodensity plots with high confidence (top) and low confidence (bottom). For each epoch, the stage-wise confidence sums to 1. In high-confidence cases, each epoch will have a dominant sleep stage with confidence close to 1. In cases with low confidence, the epoch-wise total confidence of 1 will be distributed more evenly between two or more sleep stages. The final sleep stage is the one with the highest confidence, and as more than two sleep stages may have confidence estimates above 0, the final stage's confidence may be below 0.5 if the model identifies features of multiple sleep stages within the same epoch.*

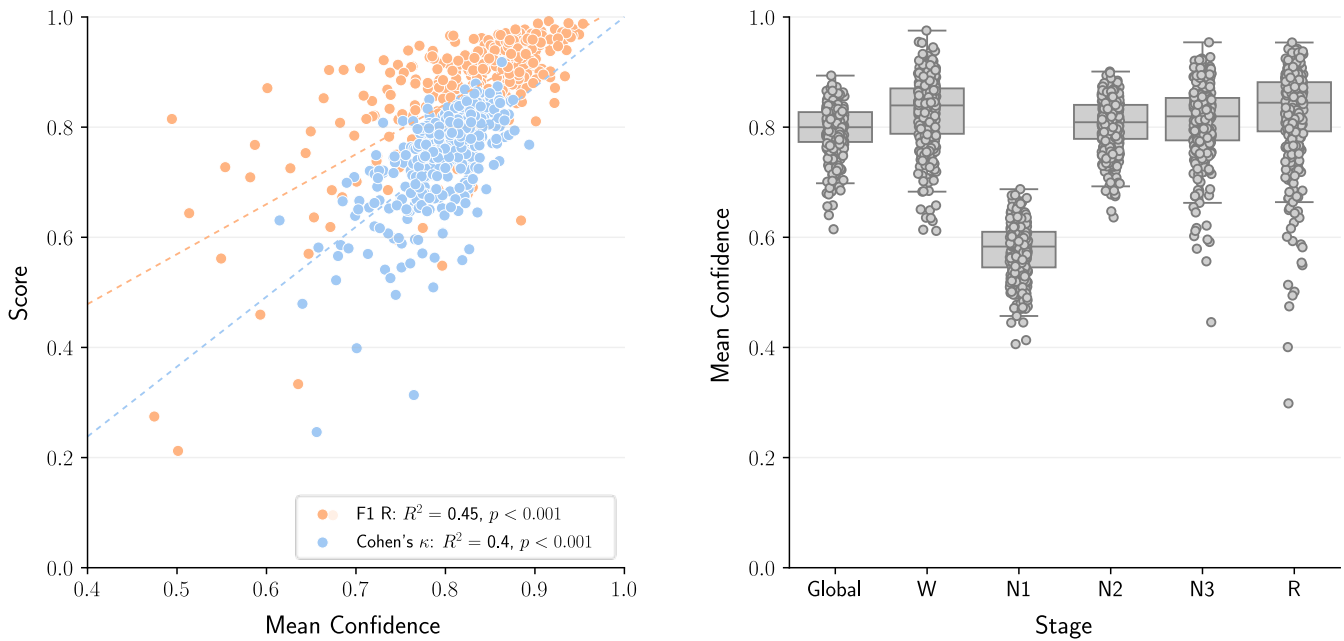

***Figure 6. Left:** Relationship between overall per-night confidence and per-night κ (blue), and the relationship between per-night REM stage confidence and REM F1 score (red). There is a clear linear relationship between the per-night confidence predicted by the model and the agreement between the model's predictions and human scoring. This relationship indicates that confidence estimates can predict agreement with human scoring, as measured by both Cohen's κ and REM F1. **Right:** Distribution of mean confidence and stage-wise confidence for the PACE and CBC dataset based on the Pretrained and the Generalized Model.*

*Impact of arousals on model confidence*

A total of 34,662 EEG arousals were manually annotated in 339 PSG recordings in the combined PACE and CBC datasets. The per-night stage-wise distributions of confidence estimates are shown in Figure S6. Only nights with at least three affected and three unaffected sleep epochs were included. Epochs manually annotated as containing arousals have significantly lower mean confidence in all stages compared with epochs unaffected by arousals (W, N1, N2, N3, R: $p < 0.001$). The mean confidence values for each stage are provided in Table S7.

**Optimized REM staging using model confidence**

The usefulness of confidence estimates was assessed by applying confidence thresholds to exclude epochs with uncertain sleep staging from further analysis. The left panel of Figure 7 shows the percentage of participants in the combined PACE and CBC datasets who retain at least 10, 30, or 50 predicted REM epochs during a single overnight polysomnogram as the confidence threshold increases. As the threshold increases, recall decreases and precision increases. With a threshold of 0.8, precision and recall were 0.96 and 0.68, respectively; raising the threshold to 0.9 yielded precision and recall of 0.98 and 0.53. The right panel of Figure 7 shows the corresponding REM sleep confusion matrices for different confidence thresholds.

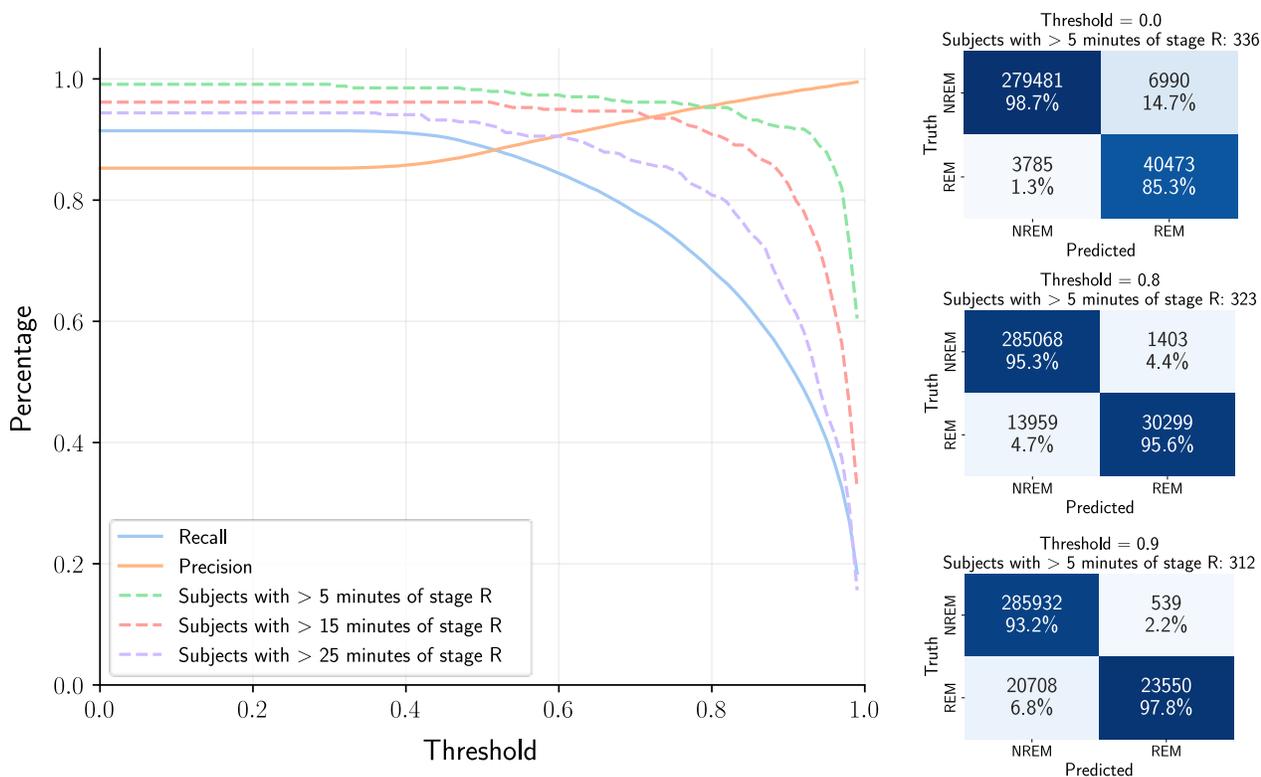

***Figure 7. Left:*** *Stage R recall and precision as a function of U-sleep confidence threshold. The curves indicate the proportion of participants retaining at least 10, 30, or 50 REM sleep epochs at each threshold.* ***Right:*** *Stage R confusion matrices for arbitrary thresholds that may be implemented to optimize subsequent clinical or automated analyses. Applying a confidence threshold of 0.8 yields a substantial improvement with 95.6% of R epochs correctly identified, compared to 85.3% without a threshold. At this threshold, 323 participants retain at least 5 minutes of high-confidence REM sleep, while 16 do not.*

## Discussion

Applying the *Pretrained Model* directly to datasets comprising patients with PD and iRBD substantially reduced performance (κ = 0.66) compared with multicenter datasets of non-neurodegenerative participants (κ = 0.82). Similarly, another study found substantially reduced accuracy with the Pretrained Model in early PD[32]. Fine-tuning the Pretrained Model on neurodegenerative datasets produced the *Generalized Model* with markedly improved agreement (κ = 0.74, $p < 0.001$). Although all models showed statistically significant differences in sleep architecture metrics, these appeared small and diminished with fine-tuning.

A previous study of interrater agreement in PD reported κ = 0.62, with N1 showing the lowest agreement. Higher age and OSAS were identified as predictors of reduced agreement[25]. Similarly, the Generalized Model showed its weakest performance in N1 and a significant negative association between κ and both age and AHI ($p < 0.001$). These parallels likely reflect intrinsic challenges and variability in human scoring, which limit the model's ability to learn generalizable rules, particularly for N1, and impose a ceiling on achievable performance. Model performance clearly exceeding human interrater agreement would raise concerns about overfitting.

U-Sleep has been shown to robustly reproduce AASM scoring in non-neurodegenerative populations[30]. However, performance may be reduced in new patient populations, such as those with neurodegenerative diseases, in which AASM rules may be applied differently. Possibly, the model may have difficulties with rules based on the absence of specific features – for example, scoring REM in epochs preceding REM (rule I3) or continuing N2 scoring without definitive N2 phenomena (K-complex or spindle, rule G4). This challenge may be especially pronounced in patients with neurodegenerative disease, in which rapid eye movements may be absent, and N2 phenomena are sparse. On average, the Generalized Model began scoring REM sleep and N2 ≈0.5 epochs earlier and continued slightly longer than human raters, resulting in increased durations of 1.23 and 0.88 epochs for REM sleep and N2, respectively. Several factors may explain this deviation. First, the model is less responsive to sleep stage transitions (Table 1, Figure S3), generally producing longer stage durations before switching, possibly due to lower sensitivity to EEG changes associated with arousals or to inconsistent manual staging of short transitions, which makes these difficult to predict. Second, the model assigns sleep stages at sample-rate resolution (128 Hz), independent of the human epoch division, which is applied post hoc. Rules that rely on epoch boundaries – for example, scoring N2 in the next epoch when a definite N2 phenomenon occurs in the last half of the current epoch – are therefore difficult for the model to reproduce, potentially biasing N2 onset earlier. Notably, the model also appears to adopt 'backward scoring' of REM sleep even when REM sleep features are sparse. Overall, these deviations in N2 and REM scoring appear minor, as both stages maintain very high accuracy.

Site-Specific Models yielded only marginal improvements, significant only in the CBC dataset. In the clinical hold-out dataset from DCSM, post hoc finetuning also failed to improve overall κ or stage-wise F1 compared to the Generalized Model. Together, these findings suggest that the Generalized Model, jointly fine-tuned on PACE and CBC, is close to the performance ceiling for U-Sleep in these cohorts. Notably, the performance either exceeds or resembles previously reported human interrater κ in PD[25,26]. A Generalized Model also offers practical advantages. Combining datasets from multiple sites and raters reduces the risk of overfitting to specific raters or local data patterns, thereby markedly improving accuracy when applied to new sleep clinics. It removes the need for fine-tuning and revalidation at each new site, and extensive fine-tuning on specific neurodegenerative data might otherwise compromise generalizability to controls or other sleep disorders – an effect not observed for the current model.

Clinically, a model must be robust both in the presence and absence of the target disease and its comorbidities. The Generalized Model demonstrated such robustness across iRBD and early-to-moderate PD, with no significant differences in per-night κ values. In the hold-out clinical dataset from DCSM, staging performance increased markedly with the Generalized Model, but a Site-Specific model showed only marginal, non-significant improvements. Notably, REM staging accuracy across all datasets did not differ between clinical subgroups. This is encouraging if the model is to be applied in automated RBD diagnostics. In contrast, N1 and N3 accuracy were significantly reduced in early-to-moderate PD compared to controls, particularly in PD$^{+RBD}$. This subgroup exhibits more widespread cortical pathology, which may lead to secondary EEG alterations and thereby complicate human and automated staging of N1 and N3[33]. However, in diagnosing RBD, non-REM sleep stages are of limited importance, whereas distinguishing W and R is critical.

In PACE and COL, control participants were recruited for iRBD screening and therefore exhibited risk factors, such as self-reported dream enactment or nocturnal movements. This control group reflects the population encountered in clinical settings and ensures comparable age and sex distributions across groups. The higher AHI in controls likely reflects obstructive sleep apnea (OSA)- related movements that mimic RBD behavior, leading to the often undiagnosed OSA patients participating in RBD screening. Although a healthier control group might have produced higher staging agreement, the present controls provide a more realistic representation of non-neurodegenerative patients encountered in clinical screening. In the future, a generalized model could be extended by incorporating datasets from other neurodegenerative diseases, such as mild cognitive impairment, DLB, and Alzheimer's disease, to further enhance applicability across the clinical spectrum.

To explore patterns underlying the lowest staging performance, all recordings with κ < 0.6 were selected for blinded re-assessment, along with 10 randomly chosen recordings with κ > 0.6. In the high-agreement group, rater 1 (R1) and the second, blinded rater (R2) showed substantial agreement across all recordings, consistent with previously reported interrater values[25]. As expected, most disagreement occurred in N1 and N3. N1 typically shows the lowest interrater agreement[19], and the discrepancies in N3 may reflect EEG abnormalities during NREM sleep in PD[23]. A similar pattern emerged when comparing the second, blinded rater (R2) with the Generalized Model (G), suggesting that when the model agrees well with one rater, comparable agreement can be expected when introducing a new rater. Conversely, for low-agreement recordings (κ < 0.6), the interrater agreement between R1 and R2 was κ = 0.49, slightly lower than the agreement of each rater with the model. This indicates that such recordings were difficult to stage for both human raters and that, in these ambiguous cases, the model may represent a balanced compromise between differing interpretations of the AASM rules. This interpretation is supported by a strong linear relationship between κ(R1 vs G) and κ(R1 vs R2) (Figure 4), implying shared sources of disagreement between model and raters.

Interestingly, the model tended to align more closely with R1 in high-agreement cases and more with R2 in challenging cases. This pattern likely reflects *regression toward the mean*: a recording with an unusually low κ when assessed by the first rater tends to regress toward the mean (a less extreme value) when reevaluated by the second rater. Consequently, recordings with unusually low model agreement tend to show slightly higher agreement with a new rater, and vice versa. Additionally, unlike human raters, the model is deterministic, providing a stable reference that may be advantageous in clinical workflows involving multiple scorers.

In an extensive interrater study comparing two European centers and the Stanford-STAGES algorithm, κ between human raters was 0.66 in sleep-clinic patients, decreasing slightly to 0.61 when comparing human scoring to the algorithm, with the lowest agreement in N1 and N3[34]. As our datasets mainly consist of elderly patients with neurodegenerative diseases and were selected based on κ, direct comparison is not possible. Nevertheless, our findings demonstrate that although the model was fine-tuned using data partly scored by R1, it does not appear overfitted, as agreement with the unseen rater (R2) remained similarly high.

Common challenges in human sleep staging - such as advanced age, OSA, and neurodegeneration - are well recognized, whereas other deviations, like poor signal quality, are rarely documented. All of these represent potential pitfalls that sleep staging models must handle. Among these challenges, the Generalized Model showed similar vulnerabilities with lower accuracy associated with higher age and AHI. However, using AHI as a predictor of automated accuracy is impractical, as AHI is derived from human scoring. Although age correlated with κ, this effect diminished after accounting for U-Sleep confidence, suggesting that EEG changes related to age are partly reflected in the model's confidence estimates. In fact, when limiting predictors to those available without manual input, the U-Sleep confidence emerged as the only significant predictor of κ (p < 0.001).

U-Sleep has previously been applied in patients with chronic insomnia and pediatric populations[35,36]. Beyond standard sleep staging, U-Sleep can be adapted to specific purposes. A recent paper used U-Sleep confidence to derive measures of global and stage-wise uncertainty and sleep-stage mixing. These novel measures were correlated with poorer motor and cognitive scores and faster motor decline, suggesting that relevant biomarkers may be derived from U-sleep models. In sleep staging, we demonstrate a strong linear relationship between confidence and staging accuracy, particularly during REM sleep, indicating that U-Sleep confidence provides a reliable measure of model trustworthiness. Clinically, confidence values could help identify recordings or segments warranting manual review, or flag data of questionable quality in retrospective datasets. Notably, two recordings showed low to moderate confidence (0.70 and 0.76) and low agreement with R1 (κ = 0.26 and κ = 0.32). At reassessment by R2, κ increased by 0.24 and 0.23, respectively, corresponding more closely with the model's confidence and illustrating how confidence can predict interrater variability. Furthermore, stage-wise confidence can be extracted at any time scale, e.g. from 30-s epochs for sleep staging to 3-s mini-epochs for RSWA quantification or even finer resolutions for microsleep analysis.

We tailored the model to optimize REM sleep staging by applying a confidence-based threshold, thereby increasing the specificity of REM sleep detection from 85% to 95.5%. Although this reduces the total amount of accepted REM sleep, 95% of subjects still have sufficient REM duration to enable quantification of RSWA[17]. Such confidence-weighted selection is not feasible in human scoring, as human raters do not quantify their confidence; consequently, all manually staged REM epochs are assumed equally valid. To address this uncertainty, RSWA quantification guidelines recommend re-staging REM sleep, focusing on archetypal REM epochs and excluding arousals, N2 phenomena, and prolonged REM absence; however, this process is time-consuming and requires specialized expertise[17]. Automated RSWA detection algorithms, such as RBDtector, have shown promising performance but rely heavily on accurate REM identification to avoid false positives[37,38]. By leveraging REM-specific U-Sleep confidence, the model provides an automated, unbiased approach for selecting archetypal REM epochs, potentially enabling fully automated RSWA quantification. Similar confidence-based adaptations could be implemented for other stages or temporal resolutions with little additional effort.

Another application of confidence estimates is the identification of uncertain epochs throughout the night, such as those affected by arousals. Comparing mean confidence values between epochs with and without manually annotated arousals revealed significantly lower confidence in arousal-affected epochs across all sleep stages ($p < 0.001$). This finding suggests that applying a confidence threshold could help automatically exclude epochs containing arousals – or, conversely, target them for review in disorders characterized by frequent arousals, such as OSAS and insomnia. Furthermore, U-Sleep confidence could serve as an informative input feature for other machine learning models.

This study has several limitations. We focused on iRBD and early-to-moderate PD, as these are clinically relevant groups for early RBD detection. Model accuracy may be lower in more advanced PD, where EEG abnormalities are more pronounced, and staging becomes increasingly complex[27]. Patients with moderate-to-late PD were included in the DCSM dataset, although clinical data were limited. Furthermore, the datasets predominantly comprise Northern European, Caucasian participants, which may restrict generalizability to more diverse populations. The interrater experiment used a relatively small sample of recordings, selected based on κ, and therefore cannot be used to infer interrater agreement across the datasets. Future studies should include broader cohorts encompassing later stages of PD and other neurodegenerative disorders, as well as more diverse populations, to assess model robustness across disease severity and demographic variation.

We demonstrate that U-Sleep achieves strong generalizability across diverse datasets, including patients with neurodegenerative disorders and elderly controls with co-morbid sleep disturbances, as well as an independent clinical dataset from an unseen site. The model can be flexibly adapted for specific purposes such as REM detection, RSWA quantification, or other stage-specific analyses. For sleep staging algorithms to become clinically relevant, cross-site and cross-population robustness is essential to avoid the need for finetuning of site-, disease-, or purpose-specific models. Generalizable models, by contrast, enable consistent evaluation of large, multi-site datasets previously scored by different raters. Ultimately, automated and standardized sleep staging represents a crucial step toward scalable, time-efficient sleep diagnostics in an era of growing clinical demand and expanding screening initiatives.

## Methods

### Datasets

*Multisite datasets for pretraining*
A publicly available, multisite dataset (PUB) comprising nearly 15,000 subjects and more than 19,000 polysomnograms from 12 sites with a diverse range of individuals, including adults and children, various sleep disorders, and healthy individuals. An overview of the datasets is provided in Table S1. All recordings include manually annotated sleep stages adhering to either the R&K or the AASM manual[39,40]. This dataset closely resembles the one used to train the U-Sleep model in Perslev et al.[28], except that all polysomnograms from DCSM were excluded to prevent the same individual from appearing in both the pretraining and hold-out testing sets.

*Neurodegenerative datasets for finetuning*
Two research datasets with PSGs, collected at the Lundbeck Foundation Parkinson's Disease Research Center at Aarhus University Hospital (PACE) and the Cologne-Bonn Cohort at the University of Cologne and University of Bonn(CBC), were included from ongoing studies. Both datasets include patients with iRBD, PD with RBD (PD[+RBD]), or PD without RBD (PD[-RBD]). Control subjects in both datasets were identified as research participants without neurodegenerative disease undergoing PSG without signs of RBD. Most controls were included in previously published studies of community-based screening for iRBD[8,9]. All polysomnograms were performed and manually annotated in accordance with the AASM manual. Subject demographics and subgroup composition are shown in Table 4.

*Clinical dataset for testing ("hold-out dataset")*

The DCSM hold-out dataset comprised patients undergoing diagnostic PSG at the Danish Centre for Sleep Medicine. This dataset is enriched with patients with iRBD and PD, some of whom have PD at an advanced stage. Controls are sleep clinic patients without evidence of RBD and without a clinical diagnosis of PD. Subject demographics and subgroup composition are shown in Table 4.

|  | Training | | | | | | Hold-out | | |
|---|---|---|---|---|---|---|---|---|---|
|  | PACE | | | CBC | | | DCSM | | |
|  | Control | iRBD | PD | Control | iRBD | PD | Control | iRBD | PD |
| Demographics | | | | | | | | | |
| N | 15 | 36 | 83 | 74 | 102 | 29 | 87 | 36 | 81 |
| Age (years) | 70.0 [65-73] | 66.5 [61-70] | 65.0 [60-70] | 62.0 [55-70] | 67.0 [61-72] | 69.0 [56-72] | 52 [36-64] | 65 [49-69] | 67 [62-71] |
| Sex (male/female) | 12/3 | 27/9 | 54/26 | 55/19 | 87/15 | 21/8 | 47/40 | 26/10 | 54/27 |
| Body mass index (BMI, kg/m$^2$) | 26.2 ± 4.3 | 25.5 ± 4.1 | 25.9 ± 4.3 | 25.7 ± 3.0 | 25.7 ± 3.1 | 25.7 ± 3.7 | 25.8 ± 5.0 | 26.3 ± 5.2 | 25.2 ± 3.4 |
| Apnea-hypopnea index (AHI, events/h) | 20.8 ± 17.7 | 10.4 ± 11.3 | 14.7 ± 14.9 | 13.5 ± 13.8 | 7.9 ± 9.7 | 7.5 ± 6.6 | 6.9 ± 7.9 | 13.4 ± 16.2 | 10.9 ± 16.2 |
| REM sleep behavior disorder (yes/no/NA) | 0/15/0 | 36/0/0 | 36/44/3 | 0/74/0 | 102/0/0 | 16/13/0 | 87/0 | 0/36 | 51/30 |
| Duration of RBD symptoms at PSG (years) |  | 5.2 ± 4.3 | 5.3 ± 5.2 |  | 12.2 ± 6.6 |  |  |  |  |
| Time since PD diagnosis PSG (years) |  |  | 1.5 ± 1.6 |  |  | 6.6 ± 4.4 |  |  |  |
| Hoehn and Yahr stage (1/2/3) |  |  | 24/48/5 |  |  |  |  |  |  |
| MDS UPDRS-III motor score (ON medication) |  |  | 19.3 ± 6.9 |  |  |  |  |  |  |
| MDS UPDRS-III motor score (OFF medication) |  | 7.4 ± 7.2 | 17.9 ± 8.0 |  | 5.3 ± 3.4 | 35.4 ± 11.1 |  |  |  |
| Levodopa equivalent daily dose (mg/day) |  |  | 226.9 ± 288.9 |  |  | 640.9 ± 431.9 |  |  |  |
| Epworth Sleepiness Scale score | 5.0 ± 2.5 | 5.4 ± 3.0 | 6.5 ± 4.1 | 7.8 ± 3.5 | 7.2 ± 3.7 | 10.0 ± 4.3 |  |  |  |
| RBD Screening Questionnaire (RBDSQ) score | 2.0 ± 1.4 | 10.1 ± 1.8 | 4.9 ± 3.1 | 7.8 ± 3.0 | 9.5 ± 1.8 | 5.4 ± 3.5 |  |  |  |
| Sniffin' Sticks 12-item identification score |  |  |  |  | 6.5 ± 2.6 |  |  |  |  |
| Sniffin' Sticks 16-item identification score | 7.7 ± 3.8 | 8.5 ± 2.8 | 7.5 ± 3.1 | 11.1 ± 2.6 |  | 7.3 ± 2.8 |  |  |  |
| Montreal Cognitive Assessment (MoCA), score | 27.5 ± 2.1 | 27.1 ± 2.6 | 27.6 ± 1.8 | 28.2 ± 0.8 | 26.9 ± 2.2 | 26.5 ± 2.4 |  |  |  |
| Sleep features | | | | | | | | | |
| Total Sleep Time (min.) | 370.8 ± 77.2 | 424.4 ± 74.3 | 394.1 ± 64.5 | 394.4 ± 70.1 | 395.4 ± 64.2 | 394.4 ± 58.3 | 411.3 ± 58.1 | 360.9 ± 76.0 | 336.6 ± 73.5 |
| Sleep Efficiency (%) | 72.9 ± 12.7 | 83.8 ± 8.4 | 76.9 ± 11.2 | 82.6 ± 10.3 | 81.9 ± 10.2 | 82.9 ± 10.0 | 87.9 ± 9.8 | 80.8 ± 10.4 | 76.6 ± 12.7 |
| Sleep Onset Latency (min.) | 9.6 ± 7.3 | 17.0 ± 15.1 | 11.1 ± 15.3 | 9.8 ± 10.9 | 11.7 ± 13.2 | 6.9 ± 6.0 | 8.0 ± 10.1 | 6.7 ± 8.2 | 5.2 ± 6.2 |
| REM Latency (min.) | 127.2 ± 69.0 | 136.7 ± 102.7 | 152.1 ± 83.7 | 104.7 ± 68.2 | 113.5 ± 77.4 | 115.6 ± 69.3 | 89.5 ± 53.4 | 137.2 ± 83.3 | 147.4 ± 105.8 |
| N1 sleep fraction (% of total sleep time) | 25.6 ± 16.6 | 16.4 ± 7.0 | 10.8 ± 7.4 | 22.0 ± 14.0 | 20.2 ± 9.9 | 15.5 ± 7.7 | 10.3 ± 8.6 | 15.3 ± 12.7 | 15.5 ± 12.4 |
| N2 sleep fraction (% of total sleep time) | 47.5 ± 12.7 | 46.8 ± 7.9 | 56.9 ± 8.1 | 48.8 ± 11.1 | 47.8 ± 9.0 | 48.1 ± 9.3 | 51.7 ± 9.7 | 50.8 ± 13.5 | 53.1 ± 17.5 |
| N3 sleep fraction (% of total sleep time) | 11.1 ± 5.8 | 17.9 ± 7.0 | 16.3 ± 8.7 | 15.2 ± 7.9 | 16.2 ± 6.6 | 15.1 ± 7.2 | 17.1 ± 10.6 | 16.2 ± 14.3 | 18.2 ± 15.7 |
| R sleep fraction (% of total sleep time) | 15.8 ± 6.6 | 18.9 ± 7.4 | 16.0 ± 5.7 | 14.1 ± 5.7 | 15.9 ± 5.5 | 21.4 ± 9.8 | 20.9 ± 6.6 | 17.6 ± 8.4 | 13.2 ± 10.0 |
| Wake After Sleep Onset (min.) | 129.3 ± 73.0 | 64.9 ± 39.1 | 105.3 ± 60.6 | 68.9 ± 46.0 | 68.8 ± 48.6 | 76.3 ± 55.1 | 51.0 ± 52.6 | 74.6 ± 44.0 | 97.8 ± 59.5 |
| Stage changes (n) | 171.7 ± 90.7 | 172.9 ± 63.7 | 144.4 ± 52.5 | 148.7 ± 48.2 | 136.7 ± 45.3 | 126.9 ± 48.4 | 119.7 ± 55.1 | 115.7 ± 49.6 | 97.2 ± 42.4 |

*Table 4* - Overview of datasets, their diagnosis composition and their demographic data, as well as their basic sleep metrics for the human sleep scorings. Data are presented as mean (SD), median [IQR], or n, as appropriate.

## Data preprocessing & Record rejection

All preprocessing was performed with our existing U-Sleep pipeline[41]. In brief, all PSG recordings were resampled to 128 Hz, scaled and outlier-clipped, and then high-pass filtered at 0.1 Hz. R&K sleep-stage annotations were converted to AASM by collapsing N3 and N4. All non-scored epochs were annotated as a separate "unknown" class, and these were not included in cost calculation or model evaluation.

For the finetuning and "hold-out" datasets, all nights were limited to their lights-off/lights-on periods during training, validation, and testing. Recordings with less than three hours of sleep (12 recordings) or with more than 20% unscored epochs (11 recordings) were rejected. All sleep epochs containing more than 50% flatline were also labelled as "unknown". Additionally, recordings with all EEG and EOG channels flatlining simultaneously for more than 10% of the recording were rejected (8 recordings). A segment of the recording was classified as a flatline if the signal had a peak-to-peak amplitude of less than $10^{-8}$ volts, evaluated over 30-second windows. After rejections, the datasets used for sleep staging consisted of 215 eligible recordings in CBC, 134 in PACE, and 204 in DCSM, as shown in Table S4. In a post hoc analysis, 10 additional recordings were excluded from CBC due to the lack of a conclusive PD and RBD diagnosis.

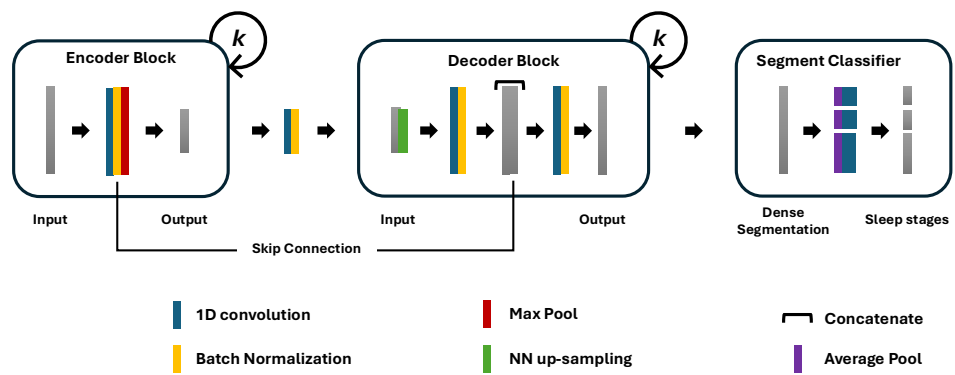

*Figure 8* - U-Sleep model architecture (adapted from Perslev et. al 2021), Open Access under Creative Commons License 4.0 http://creativecommons.org/licenses/by/4.0/])

## Sleep staging

U-Sleep is a fully convolutional neural network used for automatic sleep staging. It processes temporal data directly and was originally implemented as a 2-channel model with one EEG and one EOG channel, as described by Perslev et al.[28] U-Sleep consists of three parts: an encoder, a decoder, and a segment-classifier, as shown in Figure 8. The encoder consists of $k$ encoder layers, each comprising a 1D convolution, Batch Normalization, and a max pooling layer. The output of the encoder is a feature representation of the original signal with lower temporal resolution. This is given as input to the decoder, which consists of $k$ decoder layers. Each layer consists of an up-sampling step and two layers of 1D convolution and batch normalization. Between these two layers, a skip connection is also given, originating from its corresponding encoder block. The decoder's output is a sleep-stage segmentation with the same temporal resolution as the original signal. This is provided as input to the segment classifier, which averages across 30-second sleep epochs to produce the final sleep stage classification. The high-level overview of the training setup is shown in Figure 9. A Pretrained Model was obtained by training on large multisite datasets. The Pretrained Model was then fine-tuned on PACE and CBC, separately and combined, resulting in Site-Specific Models and a Generalized Model. The Generalized Model was tested on the clinical hold-out DCSM dataset.

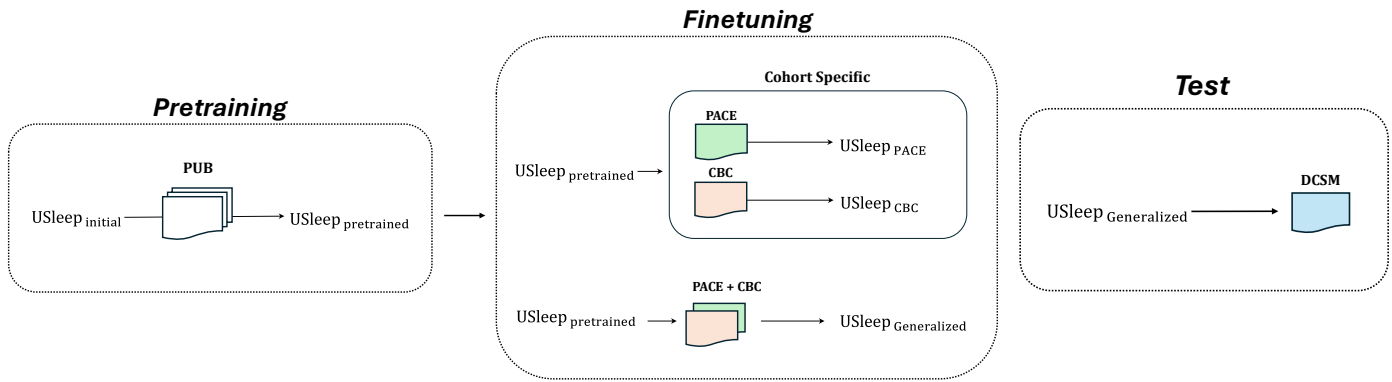

*Figure 9* - Overview of the model training setups. The Pretrained Model was fine-tuned in three ways: two specialized for specific cohorts (PACE and CBC), and the third trained on both simultaneously to obtain the Generalized Model. Finally, the Generalized Model was tested on the clinical hold-out dataset DCSM.

### Common parameters
A two-channel U-Sleep model with one random EEG and one random EOG channel from the available PSG was used for all training setups. We used an encoder/decoder depth of 12. U-Sleep has two other hyperparameters: progression factor and complexity factor, which were set to 2 and 1.67, respectively, as in Perslev et al[28]. This setup resulted in ~3.1 million trainable parameters. During training, U-Sleep was shown minibatches of 35 sleep epochs, each consisting of a single EEG and EOG channel combination. These combinations were sampled from the data using a semi-random sampling technique similar to that in Perslev et al. We used a cross-entropy loss function, the Adam Optimizer and a batch size of 64. During testing, the model was shown all possible combinations of a single EEG and EOG channel from the given PSG recording. The results from each forward pass were summed, and the majority class was determined for each sleep epoch.

### Pre-training
The PUB dataset was split into three subsets on a per-site-per-subject basis: training, validation, and test, with subject-based splits of 0.94, 0.03, and 0.03, respectively. Due to the random sampling technique, each epoch consisted of 886 minibatches. The initial learning rate was $10^{-4}$, with a decaying scheduler, patience of 40 epochs, and a down-scaling factor of 0.5. Early stopping was applied with a patience of 80 epochs, without any validation improvement.

### Fine-tuning
All fine-tuned models were evaluated using 10-fold cross-validation at the per-dataset, per-subject level. Ten subjects were set aside as validation subjects for each split. The learning rate was set at a fixed value of $10^{-5}$. Each cross-validation split ran for up to 50 epochs, with early stopping at 10 epochs. Each epoch consisted of 50 mini-batches. For testing the Generalized Model on DCSM, we combined all models from the finetuning process into an ensemble, resulting in a per-model majority-vote sleep-stage prediction.

## Sleep Staging Analysis
When possible, Cohen's κ was preferred to evaluate sleep staging performance, and F1-scores were used for comparability with previous publications and to evaluate stage-wise performance. Sleep metrics were calculated for both human and automated sleep scorings using commonly used definitions (Table S8). The metrics for each model type were calculated and compared with the human labels. Independent or pairwise permutation tests were used to determine significance, depending on the nature of the data. When evaluating Cohen's κ or F1 scores, it was always done on a per-night basis unless otherwise specified.

*REM cycle length and latency*

To investigate U-Sleep's ability to identify longer periods of REM sleep correctly, an analysis was done on start-offset and length-offset of such sleep periods. First, all REM cycles lasting for more than five epochs in both the human and automatic scoring were identified. All automatically and manually scored cycles with a single overlapping counterpart were then identified and interpreted as representing the same underlying REM sleep cycle. Finally, the starting point offset and length differences for these cycles were calculated. The Shapiro-Wilk test was used to assess normality.

**Interrater assessment inclusion criteria**

PSG recordings for the interrater assessment were chosen based on a lower-bound κ value of 0.6. An additional ten records with an initial κ value above 0.6 were selected (five from PACE, five from CBC). These were randomly sampled from five equal-sized intervals between 0.6 and 0.9 to represent the full range of performance adequately. The threshold of 0.6 was chosen such that the number of recordings included did not exceed the human resources available for rescoring.

**Hypnodensities and confidence estimates**

Hypnodensities and model confidences were derived from intermediate model outputs. The model sleep staging output can be defined as a vector $P_{sleep} = \mathbb{R}^E$, where E is the number of sleep epochs in the PSG recording. Meanwhile, model confidence can be expressed as $P_{conf} = \mathbb{R}^{E \times S}$, where S is the number of stages, and $P_{conf_{ij}}$ is the model confidence of epoch *i* being the sleep stage *j*. Furthermore, the classification of an epoch *i* is given by $P_{sleep_i} = \arg\max(p_{conf_i})$. As such, $P_{conf}$ is a matrix that describes the model's certainty and provides more insight into the decision boundaries.

*Linear Regression Model*

Three types of linear mixed regression models were fitted with Cohen's κ score as the dependent variable. Cohen's κ is bounded between -1 and 1. In our study, κ values were mainly in the range of 0.5-0.8 without approaching the boundaries, supporting the use of linear mixed effect models that assume a continuous, unbounded outcome. To account for the random variable of cohort, it was modelled as a random effect, such that all models (A, B and C) followed the same generic pattern. The fixed effects are the variables listed in Table 3.

$$Cohen's\ \kappa = fixed\ effects + (1|cohort)$$

*Validation of the linear regression model*

Model C specified in Table 3 was fitted and evaluated for all recordings in PACE and CBC using 10-fold cross-validation. The absolute residuals were averaged to find the Mean Absolute Error (MAE). Significant parameters were identified with two-tailed z-tests.

*Mean confidence and arousals*

The confidence of an epoch *i* was determined as $\max(P_{conf_i})$. The overall mean confidence of a PSG recording was therefore determined as $\frac{\sum_{i=0}^{E-1} \max(P_{conf_i})}{E}$, where E is the number of epochs in the PSG. Mean confidence for epochs affected by manually scored arousals and for epochs not affected by arousals was then calculated. An epoch was considered affected if it overlapped with the annotated arousal.

**Confidence as a threshold for REM sleep classification**

With no confidence threshold, an epoch is classified as R when $\arg\max(p_{conf_i}) = R$, with *R* being the index of the REM sleep stage. If we treat R classification as a two-class problem, we have a new matrix $P_{rem_i} \in \mathbb{R}^S$, which only contains two values – confidence of R and confidence of non-R. A confidence threshold is then introduced, such that REM sleep classification only occurs if $P_{rem_{i0}} < P_{rem_{i1}}$ and $P_{rem_{i1}} > T$, where T is a given threshold between 0 and 1.0.

## Author contributions

JS, CS, and PK conceptualized the study. JS developed and implemented the software under the supervision of KBM and PK. SS assisted in software implementation. JS and CS conducted all analyses, drafted and revised the manuscript. PK revised the manuscript. CS, NO, MO, and MS performed the diagnostic evaluations of PSG recordings. CS and NB performed the interrater study. CS, NB, DB, SR and KK, contributed to data acquisition, data management, and curation of the datasets. PB, MS, and PJJ, supervised data collection and interpretation of results, critically reviewed the manuscript, conceptualized the clinical data collection, and provided resources for their completion. All authors read and approved the final manuscript.

## Acknowledgements

This study was partly funded by Innovation Fund Denmark through the "Progression Assessment in Neurodegenerative Disorders of Ageing (PANDA)" project (Case no.: 2077-00030B).

M.S. received funding from the program "Netzwerke 2021", an initiative of the Ministry of Culture and Science of the State of Northrhine Westphalia, the Federal Ministry of Research, Technology and Space (BMFTR) under the funding code (FKZ): 01EO2107 and funding under the umbrella of the Partnership Fostering a European Research Area for Health (ERA4Health) (GA N° 101095426 of the EU Horizon Europe Research and Innovation Programme), and the European Research Council (ID 10116958). The authors thank Antonia Buchal for her assistance with data management of the CBC dataset.

P.B. is funded by research grants from Lundbeck Foundation (R359-2020-2533, R491-2024-1966) and Michael J Fox Foundation (MJFF-022856)

## Competing interests

M.S. received funding for a speaking engagement from Bial.

The remaining authors declare no financial or non-financial competing interests.

## Data availability

The datasets used in the current study are not publicly available because they contain personal health data protected by local ethics and data-protection regulations. Access requires specific approvals, including data sharing agreements. Fully anonymized, processed data may be shared on reasonable request, subject to necessary regulatory and contractual agreements.

## Code availability

The underlying code for the U-Sleep model is available at https://github.com/jesperstroem/U-Sleep-for-RBD-PD. The repository contains a detailed guide on how to set up and use the pipeline to sleep-stage various types of files containing PSG data using Python (PyTorch and PyTorch Lightning packages). The installation process is done in a few steps through a command line, and there are code examples available, such that only a few lines need to be adapted before use – meaning it requires only minimal Python experience to set up. Once set up, it is possible to use either the Pretrained Model or the Generalized model. Additionally, it is possible to customize which PSG channels are used for classification.

# Supplementary

| Dataset | Subjects | Records | Subject Composition |
|---|---|---|---|
| ABC | 49 | 132 | Adults with class II obesity and severe OSA |
| CCSHS | 515 | 515 | Children aged 8-11 years |
| CFS | 730 | 730 | 2284 subjects across 361 families |
| CHAT | 1 232 | 1 639 | Children aged 5-10 years with OSA |
| HPAP | 247 | 247 | Adults with moderate to severe OSA |
| MESA | 2 055 | 2 055 | Multi-ethnic group aged 45-84 years |
| MROS | 2 903 | 3 921 | Men aged 65 or older |
| PHYS | 919 | 919 | Patients from MGH sleep center |
| SEDF-SC | 76 | 145 | Healthy adults aged 25-101 |
| SEDF-ST | 21 | 36 | Adults under the influence of temazepam |
| SHHS | 5 797 | 8 444 | Adults aged 40 or older |
| SOF | 453 | 453 | Women aged 65-89 years |
| Total | 14997 | 19236 | |

*Table S1* - Overview of the multi-site dataset (PUB) used to train the base U-Sleep model

| | | | F1 | | | | | |
|---|---|---|---|---|---|---|---|---|
| Cohort | Accuracy | Cohen's κ | Wake | N1 | N2 | N3 | REM | Macro |
| ABC | 0.82 ± 0.02 | 0.76 ± 0.03 | 0.86 ± 0.02 | 0.60 ± 0.10 | 0.84 ± 0.02 | 0.75 ± 0.07 | 0.91 ± 0.05 | 0.79 ± 0.00 |
| CHAT | 0.89 ± 0.04 | 0.85 ± 0.05 | 0.95 ± 0.03 | 0.64 ± 0.09 | 0.87 ± 0.05 | 0.91 ± 0.05 | 0.90 ± 0.07 | 0.85 ± 0.04 |
| CCSHS | 0.90 ± 0.04 | 0.87 ± 0.06 | 0.97 ± 0.02 | 0.54 ± 0.19 | 0.89 ± 0.05 | 0.86 ± 0.08 | 0.91 ± 0.04 | 0.84 ± 0.06 |
| CFS | 0.87 ± 0.08 | 0.80 ± 0.11 | 0.94 ± 0.07 | 0.41 ± 0.17 | 0.85 ± 0.07 | 0.69 ± 0.24 | 0.83 ± 0.19 | 0.74 ± 0.10 |
| HOMEPAP | 0.82 ± 0.07 | 0.75 ± 0.09 | 0.88 ± 0.09 | 0.49 ± 0.17 | 0.83 ± 0.08 | 0.73 ± 0.24 | 0.89 ± 0.05 | 0.76 ± 0.09 |
| MESA | 0.88 ± 0.06 | 0.82 ± 0.09 | 0.95 ± 0.05 | 0.57 ± 0.13 | 0.87 ± 0.06 | 0.55 ± 0.31 | 0.89 ± 0.14 | 0.77 ± 0.09 |
| MROS | 0.89 ± 0.04 | 0.83 ± 0.07 | 0.95 ± 0.06 | 0.38 ± 0.16 | 0.87 ± 0.06 | 0.57 ± 0.27 | 0.86 ± 0.16 | 0.73 ± 0.09 |
| SEDF-SC | 0.88 ± 0.04 | 0.78 ± 0.07 | 0.98 ± 0.00 | 0.38 ± 0.12 | 0.69 ± 0.15 | 0.57 ± 0.22 | 0.80 ± 0.17 | 0.69 ± 0.06 |
| SEDF-ST | 0.86 ± 0.03 | 0.81 ± 0.04 | 0.86 ± 0.01 | 0.60 ± 0.18 | 0.87 ± 0.02 | 0.89 ± 0.01 | 0.89 ± 0.03 | 0.82 ± 0.05 |
| SHHS | 0.87 ± 0.07 | 0.81 ± 0.11 | 0.93 ± 0.08 | 0.47 ± 0.20 | 0.87 ± 0.08 | 0.71 ± 0.25 | 0.87 ± 0.19 | 0.77 ± 0.12 |
| SOF | 0.88 ± 0.05 | 0.82 ± 0.07 | 0.94 ± 0.04 | 0.38 ± 0.16 | 0.86 ± 0.05 | 0.80 ± 0.12 | 0.90 ± 0.06 | 0.78 ± 0.06 |
| PHYS | 0.83 ± 0.05 | 0.76 ± 0.07 | 0.79 ± 0.11 | 0.61 ± 0.12 | 0.86 ± 0.07 | 0.71 ± 0.26 | 0.86 ± 0.13 | 0.77 ± 0.08 |
| **Mean** | **0.88 ± 0.06** | **0.82 ± 0.09** | **0.93 ± 0.08** | **0.48 ± 0.19** | **0.87 ± 0.07** | **0.68 ± 0.27** | **0.87 ± 0.16** | **0.77 ± 0.10** |

*Table S2* – Mean and standard deviation of per-night Cohen's κ and per-night F1 scores when testing the Pretrained Model on the large multi-site dataset

| Cohort | Our model | Perslev et al. |
|---|---|---|
| ABC | 0.81 | 0.77 |
| CHAT | 0.86 | 0.85 |
| CCSHS | 0.84 | 0.85 |
| CFS | 0.77 | 0.82 |
| HOMEPAP | 0.78 | 0.78 |
| MESA | 0.81 | 0.79 |
| MROS | 0.76 | 0.77 |
| SEDF-SC | 0.70 | 0.79 |
| SEDF-ST | 0.83 | 0.76 |
| SHHS | 0.80 | 0.80 |
| SOF | 0.79 | 0.78 |
| PHYS | 0.81 | 0.79 |
| **Mean** | **0.80** | **0.80** |

*Table S3 – Per-dataset Macro F1 scores for our Pretrained Model and the model reported by Perslev et al.*

| | | Rejection reasons | | | | | |
|---|---|---|---|---|---|---|---|
| Site | Total Recordings | Less than 3 hours of sleep | More than 20% unscored epochs | More than 10% flatline in all channels simultaneously | Remaining for training and test | Inconclusive PD and RBD diagnosis | Used in analysis and results |
| PACE | 136 | 1 | 1 | 0 | 134 | 0 | 134 |
| COL | 231 | 3 | 5 | 8 | 215 | 10 | 205 |
| DCSM | 217 | 8 | 5 | 0 | 204 | 0 | 204 |
| Total | 584 | 12 | 11 | 8 | 553 | 10 | 543 |

*Table S4 - Overview of all recordings and the number of rejected recordings across datasets, as well as the reasons for rejection. Ten recordings from CBC were left out of the analysis and results due to an inconclusive PD and RBD diagnosis.*

| Cohort | Pretrained | Generalized | Site Specific |
|---|---|---|---|
| PACE | 0.65 ± 0.11<br>0.67 [0.59-0.72] | **0.72 ± 0.09*** <br>**0.73 [0.67-0.79]*** | 0.72 ± 0.09<br>0.74 [0.67-0.79] |
| CBC | 0.67 ± 0.09<br>0.69 [0.62-0.74] | **0.76 ± 0.08*** <br>**0.76 [0.72-0.81]*** | 0.76 ± 0.08<br>**0.77 [0.73-0.81]*** |
| DCSM | 0.6 ± 0.17<br>0.64 [0.53-0.72] | **0.64 ± 0.18*** <br>**0.69 [0.59-0.76]*** | 0.66 ± 0.18<br>0.7 [0.59-0.78] |

*Table S5 – Mean ± SD and Median [IQR] of per-night Cohen's κ for each model type and each cohort. * indicates a significant improvement from the row and column directly to the left (p < 0.05).*

| | Cohen's κ | | | F1 scores – Generalized | | | | | |
|---|---|---|---|---|---|---|---|---|---|
| Subgroup | Pretrained | Generalized | Site Specific | W | N1 | N2 | N3 | REM | Macro |
| Control | 0.67±0.09 | 0.75±0.09 | 0.75±0.09 | 0.82±0.11 | 0.59±0.12 | 0.84±0.09 | 0.81±0.15 | 0.87±0.12 | 0.78±0.07 |
| iRBD | 0.65±0.10 | 0.74±0.08 | 0.74±0.08 | 0.80±0.11 | 0.53±0.13 | 0.84±0.06 | 0.83±0.13 | 0.86±0.11 | 0.77±0.06 |
| PD$^{-RBD}$ | 0.69±0.10 | 0.75±0.09 | 0.76±0.09 | 0.85±0.10 | 0.46±0.15 | 0.84±0.08 | 0.78±0.16 | 0.90±0.08 | 0.77±0.07 |
| PD$^{+RBD}$ | 0.66±0.10 | 0.72±0.09 | 0.73±0.08 | 0.82±0.12 | 0.44±0.15 | 0.84±0.06 | 0.75±0.12 | 0.85±0.10 | 0.74±0.07 |

*Table S6 – Mean and standard deviation of Cohen's κ values across models and clinical subgroups for the PACE and CBC cohorts. Additionally, the mean and standard deviation of F1 scores for the Generalized Model across clinical subgroups.*

| Affected by arousal | Wake Confidence | N1 Confidence | N2 Confidence | N3 Confidence | REM Confidence | Global Confidence |
|---|---|---|---|---|---|---|
| No | 0.83±0.06 | 0.58±0.05 | 0.81±0.05 | 0.81±0.06 | 0.83±0.07 | 0.77±0.12 |
| Yes | 0.77±0.09 | 0.57±0.05 | 0.74±0.07 | 0.70±0.08 | 0.75±0.10 | 0.71±0.11 |

*Table S7 – Per-night mean confidence scores calculated for the epochs with and without arousals separately. There is a decline in confidence across all sleep stages when epochs are affected by arousal.*

| Metric | Abbrev. | Description |
|---|---|---|
| Total Sleep Time | TST | The total amount of sleep in minutes. |
| Sleep Efficiency | SEFF | The fraction of sleep epochs compared to the total number of epochs. |
| Sleep Onset Latency | SOL | Minutes until the first sleep epoch. |
| REM Latency | REMLAT | Minutes until the first R epoch |
| N1 fraction | PN1 | The fraction of time spent in N1 compared to the Total Sleep Time |
| N2 fraction | PN2 | The fraction of time spent in N2 compared to the Total Sleep Time |
| N3 fraction | PN3 | The fraction of time spent in N3 compared to the Total Sleep Time |
| R fraction | PREM | The fraction of time spent in R compared to the Total Sleep Time |
| Wake After Sleep Onset | WASO | Minutes spent in W after the first sleep epoch. |
| Stage changes | STAGEC | Total number of stage changes throughout the entire night. |

*Table S8 - Overview of sleep metrics calculated to review the performance of U-Sleep. All metrics are from the AASM manual except for Stage Changes.*

| | | | F1 | | | | |
|---|---|---|---|---|---|---|---|
| Model | Accuracy | Cohen's κ | Wake | Light Sleep | Deep Sleep | REM | Average |
| Pretrained | 0.83±0.06 | 0.72±0.10 | 0.74±0.13 | 0.86±0.05 | 0.73±0.20 | 0.79±0.19 | 0.78±0.09 |
| Generalized | 0.87±0.05 | 0.79±0.09 | 0.82±0.11 | 0.88±0.05 | 0.79±0.17 | 0.86±0.14 | 0.84±0.08 |
| Site Specific | 0.88±0.05 | 0.79±0.09 | 0.82±0.11 | 0.88±0.05 | 0.79±0.17 | 0.86±0.14 | 0.84±0.08 |

*Table S9 - Overall and stage-wise model performance of all three model types in a 4-stage classification setup. N1 and N2 have been merged to form the "Light Sleep" stage, and "Deep Sleep" represents N3.*

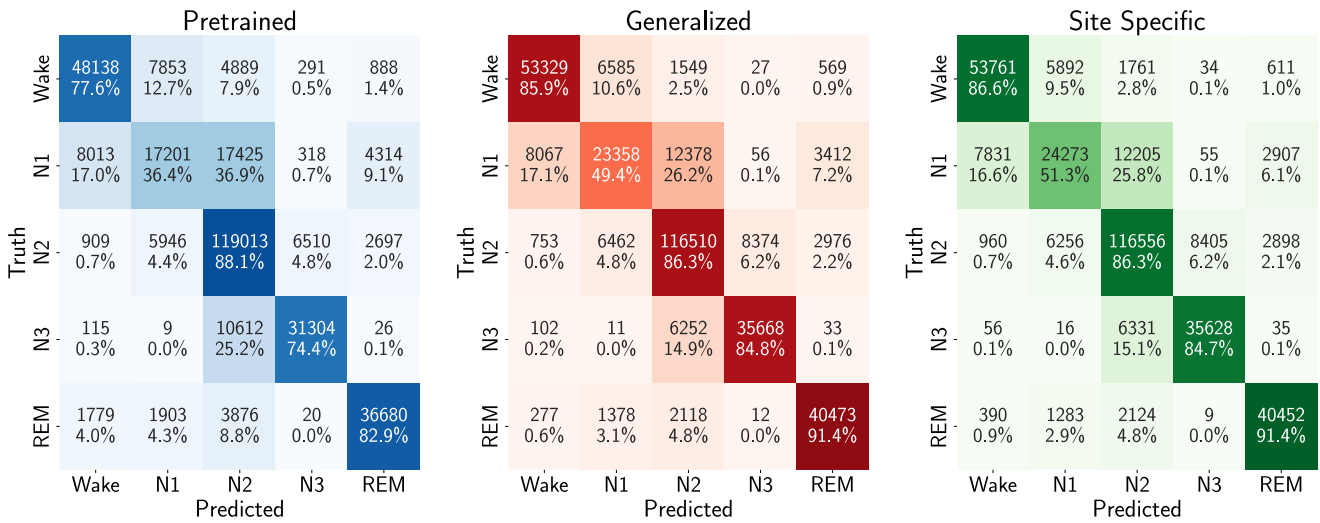

*Figure S1* - Confusion matrices for the Pretrained, Generalized and Site-Specific Models.

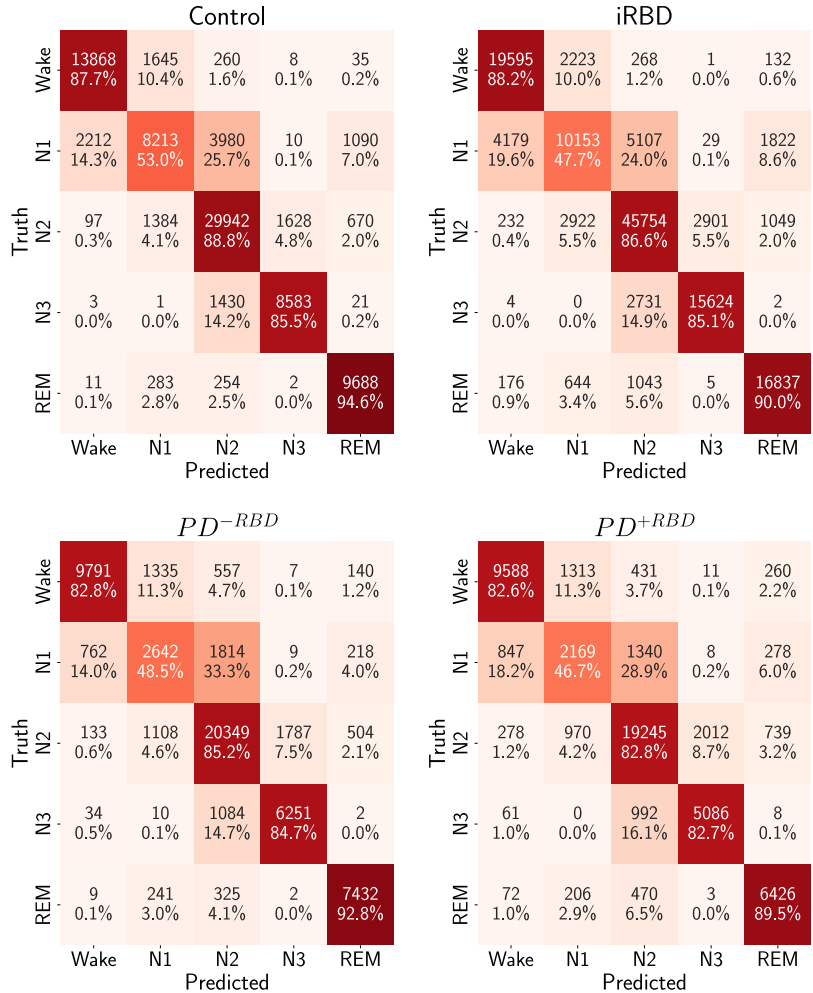

*Figure S2* - Confusion matrices for the Generalized Model across the four clinical subgroups.

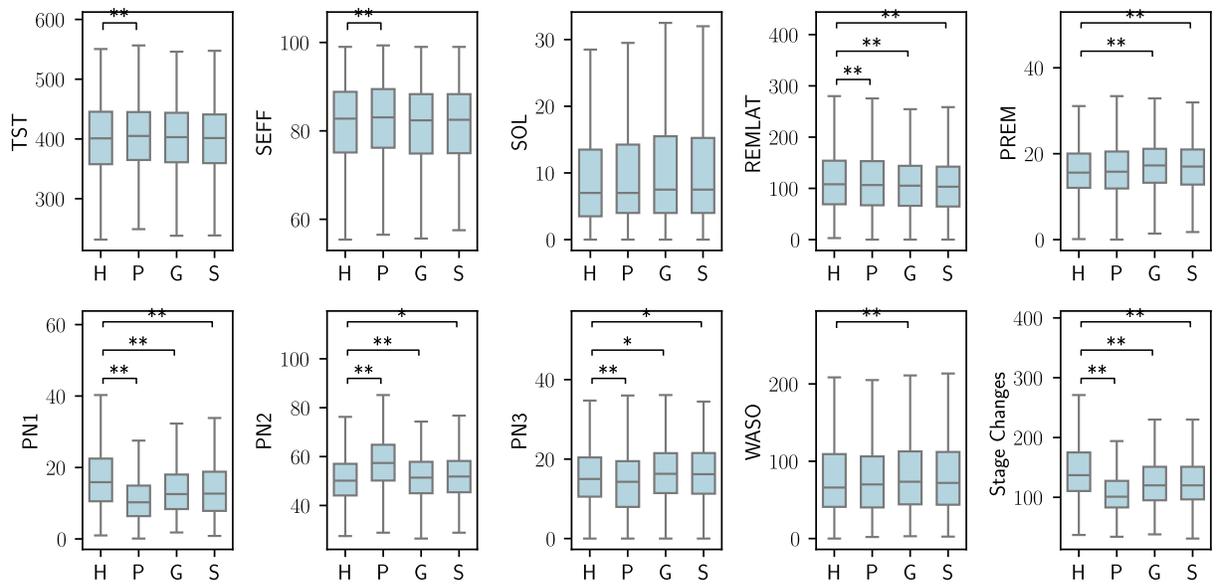

***Figure S3*** - *Sleep architecture measures per record for each model compared to human scorings. H = Human scorer, P = Pretrained Model, G = Generalized Model, S = Site Specific Model. Stars indicate a significant difference from the manual scorings. \*\* indicates a p-value < 0.01, while \* indicates a p-value < 0.05.*

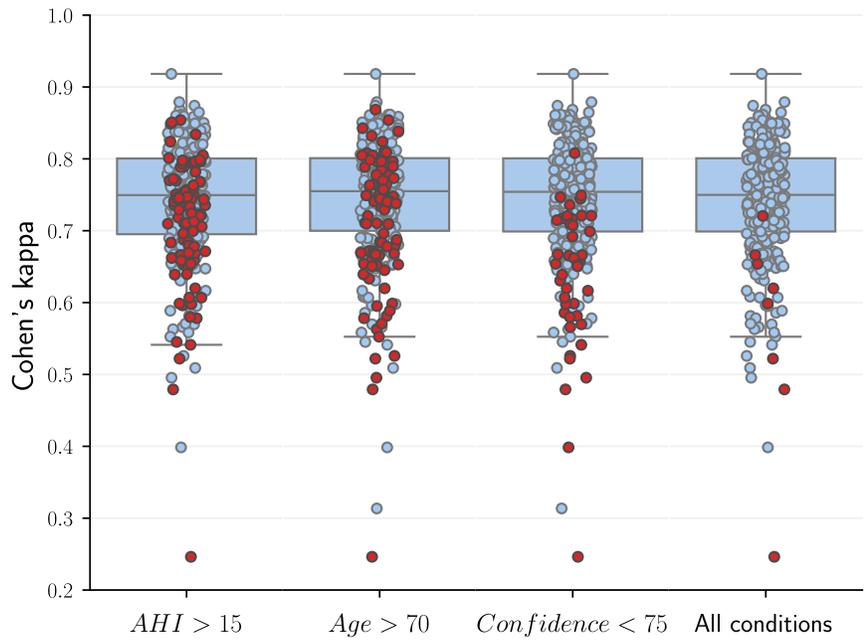

***Figure S4*** - *Distribution of mean subject-wise confidence scores for the significant variables included in the linear regression models (A, B and C). The red dots indicate individuals crossing the threshold for the particular variable. The recordings marked with red dots are considered at risk of being imprecisely staged by the model, as we observed linear relationships between Age and κ, AHI and κ, and per-night confidence and K.*

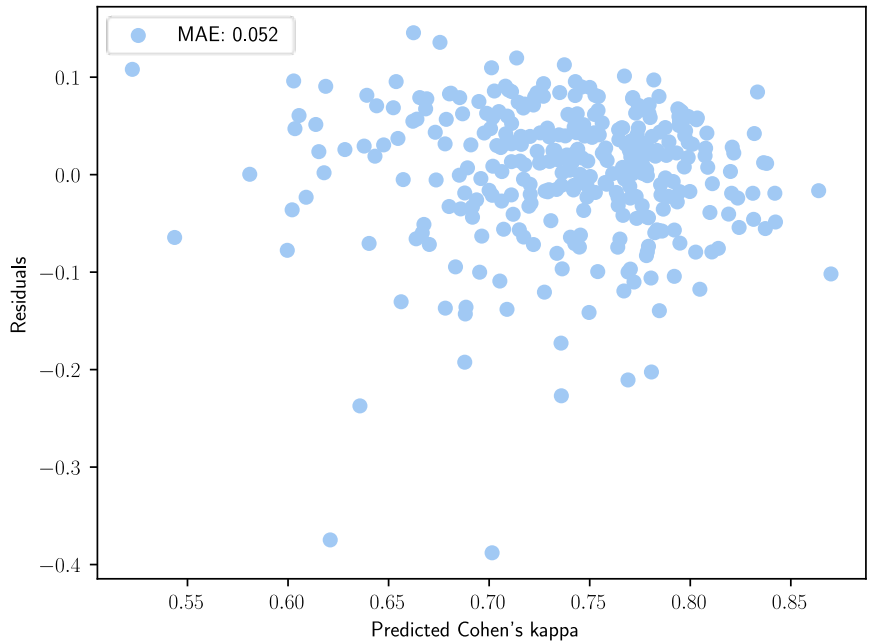

***Figure S5*** *– Residuals as a function of predicted Cohen's κ for model C, when fitting and testing on PACE and CBC with 10-fold cross-validation. The Mean Absolute Error (MAE) is 0.052, indicating that, on average, we can expect an error of 0.052 κ from the fitted model on new, unseen data.*

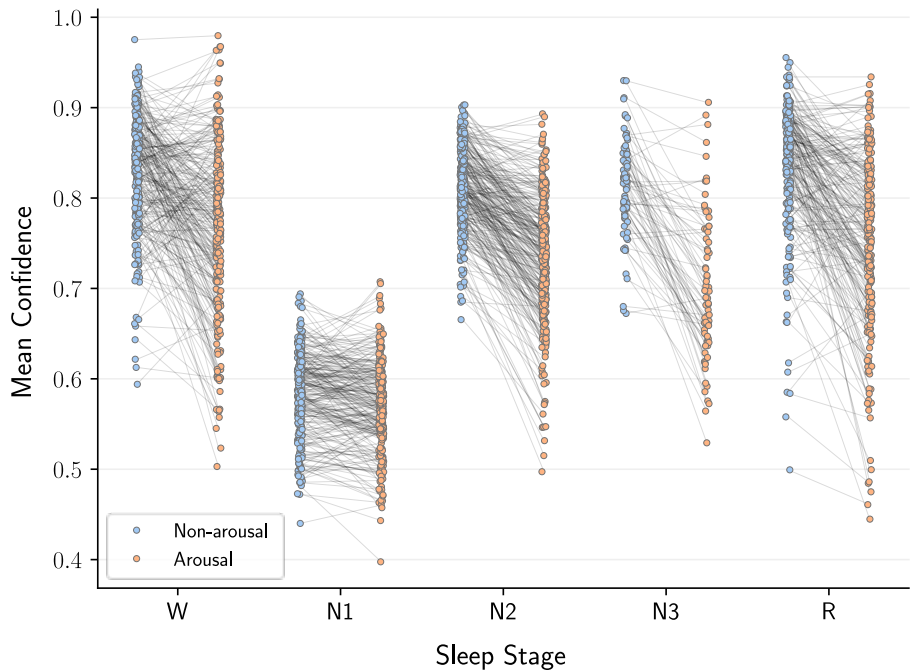

***Figure S6*** *- Model confidence across sleep stages for epochs affected by arousals compared to unaffected epochs. Epochs manually annotated as affected by arousals have significantly lower mean confidence across all sleep stages ($p < 0.05$). Each paired data point represents one stage from one PSG recording.*